\documentclass{article}

\usepackage{templatepackage/PRIMEarxiv}

\usepackage[utf8]{inputenc} \usepackage[T1]{fontenc}    \usepackage{hyperref}       \usepackage{url}            \usepackage{booktabs}       \usepackage{amsfonts}       \usepackage{nicefrac}       \usepackage{microtype}      \usepackage{lipsum}
\usepackage{fancyhdr}       \usepackage{graphicx}       \graphicspath{{media/}}     

\usepackage[numbers]{natbib}
\usepackage[affil-it]{authblk}
\usepackage{physics}
\usepackage{amsfonts}
\ifcsname proof\endcsname
\else
\usepackage{amsthm}
\fi
\usepackage{amssymb}
\usepackage{cleveref}
\usepackage{hhline}
\usepackage{multirow}
\usepackage{enumitem}
\ifcsname citet\endcsname
\else
  \usepackage{natbib}
\fi
\ifcsname cellcolor\endcsname
\else
  \usepackage[table]{xcolor}
\fi
\ifcsname mathclap\endcsname
\else
  \usepackage{mathtools}
\fi
\ifcsname tcolorbox\endcsname
\else
  \usepackage{tcolorbox}
\fi
\usepackage{subcaption}
\usepackage{bm}
\usepackage{xfrac}
\ifcsname makecell\endcsname
\else
  \usepackage{makecell}
\fi
\usepackage{listings} \newcommand{\opn}[1]{\operatorname{#1}}
\DeclarePairedDelimiterXPP\inner[2]{}{\langle}{\rangle}{}{#1,#2}
\allowdisplaybreaks
\makeatletter
\newcommand{\myvspace}{\@ifstar\myvspacestar\myvspacenostar}
\newcommand{\myvspacenostar}[1]{}
\newcommand{\myvspacestar}[1]{}
\newcommand{\crvspace}{\@ifstar\crvspacestar\crvspacenostar}
\newcommand{\crvspacenostar}[1]{\vspace{#1}}
\newcommand{\crvspacestar}[1]{\vspace*{#1}}
\makeatother \ifcsname theorem\endcsname
\else
    \newtheorem{theorem}{Theorem}
\fi
\ifcsname definition\endcsname
\else
    \newtheorem{definition}[theorem]{Definition}
\fi
\ifcsname lemma\endcsname
\else
    \newtheorem{lemma}[theorem]{Lemma}
\fi
\ifcsname remark\endcsname
\else
    \newtheorem{remark}[theorem]{Remark}
\fi
 \ifcsname algorithmic\endcsname
\else
  \usepackage{algorithm}
  \usepackage{algorithmic}
\fi

  \usepackage{tikz}
\usetikzlibrary{arrows.meta, positioning}
\usetikzlibrary{calc,math}
\ifcsname usepgfplotslibrary\endcsname
\else
  \usepackage{pgfplots}
\fi
\usepgfplotslibrary{fillbetween}
\pgfplotsset{compat=1.18}  
\title{Towards Optimal Multi-draft Speculative Decoding
}

\author[1,2,$\ast$]{\textbf{Zhengmian Hu}}
\author[1,$\ast$]{\textbf{Tong Zheng}}
\author[2, 3]{\textbf{Vignesh Viswanathan}}
\author[1]{\textbf{Ziyi Chen}}
\author[2]{\textbf{Ryan A. Rossi}}
\author[1]{\textbf{Yihan Wu}}
\author[1,4]{\textbf{Dinesh Manocha}}
\author[1]{\textbf{Heng Huang}}

\affil[1]{Department of Computer Science, University of Maryland, College Park, MD, USA}
\affil[2]{Adobe Research, San Jose, CA, USA}
\affil[3]{Manning College of Information \& Computer Sciences, University of Massachusetts Amherst, MA, USA}
\affil[4]{Department of Electrical and Computer Engineering, University of Maryland, College Park, MD, USA}

\begin{document}
\maketitle
\begin{abstract}
Large Language Models (LLMs) have become an indispensable part of natural language processing tasks. However, autoregressive sampling has become an efficiency bottleneck. Multi-Draft Speculative Decoding (MDSD) is a recent approach where, when generating each token, a small draft model generates multiple drafts, and the target LLM verifies them in parallel, ensuring that the final output conforms to the target model distribution. The two main design choices in MDSD are the draft sampling method and the verification algorithm. For a fixed draft sampling method, the optimal acceptance rate is a solution to an optimal transport problem, but the complexity of this problem makes it difficult to solve for the optimal acceptance rate and measure the gap between existing verification algorithms and the theoretical upper bound. This paper discusses the dual of the optimal transport problem, providing a way to efficiently compute the optimal acceptance rate for sampling with replacement and conditional Poisson sampling (CPS). For the first time, we measure the theoretical upper bound of MDSD efficiency for vocabulary sizes in the thousands and quantify the gap between existing verification algorithms and this bound. We also compare different draft sampling methods based on their optimal acceptance rates. Our results show that the draft sampling method strongly influences the optimal acceptance rate, with CPS outperforming sampling with replacement. Additionally, existing verification algorithms do not reach the theoretical upper bound for with replacement sampling. Our findings suggest that carefully designed draft sampling methods can potentially improve the optimal acceptance rate and enable the development of verification algorithms that closely match the theoretical upper bound.
\end{abstract} 
\renewcommand{\thefootnote}{\fnsymbol{footnote}}
\footnotetext[1]{Co-first authors.}

\section{Introduction}\label{se:introduction}
Autoregressive language models have achieved state-of-the-art results in various language tasks \citep{NEURIPS2020_1457c0d6,touvron2023llama}, including chatbots \citep{luo2022critical} and code generation \citep{chen2021evaluating}. These models generate outputs by predicting the next token sequentially. However, this autoregressive decoding process leads to significant computational resource requirements and high latency, posing challenges for user experience and limiting potential applications.

Speculative decoding \citep{leviathan2023fast,chen2023accelerating} has been proposed to address the high inference cost issue. The method uses a small, fast draft model to generate candidate results, which are then verified and corrected by a large, accurate target model to maintain the model output distribution. Compared to other acceleration methods, such as knowledge distillation, model quantization, and model pruning, speculative decoding has the advantage of significantly reducing inference latency without sacrificing quality of the generated content.

Multi-Draft Speculative Decoding (MDSD) \citep{miao2024specinfer,cai2024medusa,li2024eagle,spector2023accelerating} is a recent advancement in speculative decoding. When generating each token, the small draft model generates multiple draft tokens instead of a single one, as in vanilla speculative decoding. The target LLM verifies these tokens in parallel, ensuring that the final output aligns with the target model's distribution while achieving a higher overall acceptance rate than vanilla speculative decoding, as the multiple drafts provide better coverage of the target model's possible outputs.

MDSD algorithms have two main design choices: (1) The draft sampling method. Common approaches include sampling with replacement, where each token is independently sampled from the draft model output distribution, and successive sampling without replacement, where the probability of selecting a token is updated after each draw to exclude previously selected tokens. (2) The verification algorithm design. Examples include Recursive Rejection Sampling (RRS) \citep{yang2024multi,jeon2024recursive}, which sequentially verifies the draft tokens, and K-SEQ \citep{sun2024spectr}, which is designed to improve acceptance rate for sampling with replacement.

The acceptance rate, a measure of MDSD algorithm performance, also depends on these two design choices. Any verification algorithm that guarantees the final output aligns with the target model distribution can be viewed as a transport from the draft tokens' distribution to the target model's distribution. For a fixed draft sampling method, the optimal verification algorithm is a solution to an optimal transport problem \citep{sun2024spectr}, corresponding to an optimal acceptance rate.

However, the complexity of this optimal transport problem, with the number of variables and constraints growing exponentially with the number of draft tokens, makes it difficult to find efficient solutions. This difficulty has led to two open questions:

(1) For modern LLMs, where the vocabulary size is typically in the thousands, the optimal acceptance rate has never been computed, to the best of our knowledge. Simple linear program (LP) solvers can only compute the optimal transport for small toy models, making it challenging to measure the optimal acceptance rate in practical scenarios.

(2) Although it is widely known that existing verification algorithms are only approximate solutions to the optimal transport problem, the gap between their performance and the theoretical upper bound has never been quantified with respect to real text distribution. Without knowing the optimal acceptance rate, it is difficult to assess how suboptimal these algorithms are.

This paper addresses these two open questions. Our contributions include:
\crvspace{-5pt}
\begin{itemize}[leftmargin=*]
\item We transform the problem of solving the optimal acceptance rate corresponding to the optimal transport into a subset selection problem by considering the dual of the problem and then applying total unimodularity. This provides a novel perspective for understanding the efficiency of MDSD.
\item For certain special cases, we propose efficient methods to solve the subset selection problem by noticing convexity-like structures in the set function. This includes sampling with replacement and conditional Poisson sampling. For the first time, we provide a practical method to compute the theoretical acceptance rate upper bound of MDSD for a draft distribution.
\item For the first time, we measure the theoretical upper bound of MDSD efficiency on real text, and the gap of existing verification algorithms. We compare different draft sampling methods through their optimal acceptance rates and observe that CPS outperforms sampling with replacement. We evaluate existing verification algorithms, including K-SEQ for with replacement and RRS for without replacement and with replacement sampling, and find that the with-replacement methods still have significant gaps from the theoretical upper bound.
\item We propose a novel draft sampling method that greedily selects high-probability drafts, with only the last draft being random. In some cases, it achieves an even higher optimal acceptance rate than CPS. We also propose a corresponding verification algorithm that perfectly reaches the theoretical acceptance rate upper bound.
\end{itemize}

 \crvspace{-10pt}
\section{Preliminaries}\label{se:preliminary}
\crvspace{-10pt}
\subsection{Speculative Decoding for Accelerating LLM Inference}\label{sec:prelim_basic_sps}
\crvspace{-10pt}
Let $\Sigma$ denote the vocabulary set. We have a target model $P_{\opn{target}}(\cdot|x_1,x_2,...,x_m)$, which is a probabilistic model that predicts the probability of the next word. Our goal is to sample from this model as the output.

The process of single step Multi-Draft Speculative Decoding is as follows:
\begin{enumerate}[leftmargin=*,noitemsep=0mm,topsep=-2pt]
\item For a draft model $P_{\opn{draft}}(\cdot|x_1,x_2,...,x_m)$, sample $n$ draft tokens $\widehat{x}^{(1)},\dots\widehat{x}^{(n)}$.
\item Compute the probabilities of the target model in parallel:
$P_{\opn{target}}(\cdot|x_1,x_2,...,x_m)$, $P_{\opn{target}}(\cdot|x_1,x_2,...,x_m,\widehat{x}^{(1)})$, ..., $P_{\opn{target}}(\cdot|x_1,x_2,...,x_m,\widehat{x}^{(n)})$. Due to parallel computation, this step is not much slower than computing $P_{\opn{target}}(\cdot|x_1,x_2,...,x_m)$ alone.
\item Run the verification algorithm $x_{m+1}\sim P_{\opn{verify}}(\cdot|\widehat{x}^{(1)},\dots\widehat{x}^{(n)})$.
\item If accepted, for some draft $\widehat{x}^{(i)}$, we have $x_{m+1}=\widehat{x}^{(i)}$. In this case, we can perform another sampling step $x_{m+2}\sim P_{\opn{target}}(\cdot|x_1,x_2,...,x_m,\widehat{x}^{(i)})$, generating two tokens in one step and achieving acceleration.
\end{enumerate}

Speculative Decoding can generate multiple steps, with multiple drafts at each step and all drafts forming a tree. However, we only consider the single-step case in this paper. For following analysis, we use $p(\cdot)=P_{\opn{target}}(\cdot|x_1,x_2,...,x_m)$ to denote the target distribution and  $p_{\opn{draft}}$ for distribution of draft tokens.

\myvspace{-10pt}
\crvspace{-10pt}
\subsection{Speculative Decoding with a Single Draft Token}
\myvspace{-5pt}
\crvspace{-5pt}
Informally, the verification algorithm depends on two distributions $p$ and $p_{\opn{draft}}$, and one draft token $j\sim p_{\opn{draft}}$. The goal is to output $i\sim p$ such that the objective $\max P(i=j)$ is achieved, that is to maximize the probability of random variable $i$ to be the same as random variable $j$.

More formally, given $p\in \Delta_\Sigma$ and $p_{\opn{draft}}\in \Delta_\Sigma$ representing two probability distributions over the space $\Sigma$, we seek a joint distribution $\pi\in\Pi(p,p_{\opn{draft}})$ such that the marginal distributions are $p$ and $p_{\opn{draft}}$, respectively, and the objective $\max \sum_{i\in\Sigma} \pi(i,i)$ is maximized. This forms a optimal transport problem. The optimal transport is denoted as $\pi^\ast_{p,p_{\opn{draft}}}\in\Pi(p,p_{\opn{draft}})$, and the optimal objective function value is $\alpha^\ast(p,p_{\opn{draft}})= \sum_{i\in\Sigma} \pi^\ast_{p,p_{\opn{draft}}}(i,i)$.

The problem can be formulated as an LP by representing the joint distribution as a matrix:
\myvspace{-4pt}
\begin{equation}
\max_{C \in \mathbb{R}^{\Sigma \times \Sigma}}
\sum_{i \in \Sigma} C_{i,i}
~~
\text{s.t.~}
\sum_{j \in \Sigma} C_{i,j} = p(i), \ \sum_{i \in \Sigma} C_{i,j} = p_{\text{draft}}(j), \ C_{i,j} \geq 0 \  \forall i,j \in \Sigma.
\end{equation}
\par\myvspace{-10pt}
\crvspace{-8pt}
The optimal transport has the following closed-form expression
\myvspace{-3pt}
\crvspace{-3pt}
\begin{equation}
\pi^\ast_{p,p_{\opn{draft}}}(i,j)=C^\ast_{i,j}=\begin{cases}\min(p(i),p_{\opn{draft}}(i))&i=j\\\frac{(p(i)-p_{\opn{draft}}(i))_+(p_{\opn{draft}}(j)-p(j))_+}{\sum_{z\in\Sigma}(p_{\opn{draft}}(z)-p(z))_+}&i\neq j\end{cases}
~,
\end{equation}
and the optimal objective function value is
\myvspace{-3pt}
\crvspace{-3pt}
\begin{equation}
\alpha^\ast(p,p_{\opn{draft}})=\sum_{i\in\Sigma}\min(p(i),p_{\opn{draft}}(i))
~.
\end{equation}
\par\myvspace{-10pt}
The conditional distribution of $i$ given $j$ when $(i,j)\sim\pi$ is denoted as $\pi_{}(\cdot|j)\in G(\Sigma)$, where
\myvspace{-3pt}
\begin{equation}
\pi_{}(i|j)=\frac{\pi_{i,j}}{\sum_{i\in\Sigma}\pi_{i,j}}=\frac{\pi_{i,j}}{p_{\opn{draft}}(j)}
~.
\end{equation}
\par\myvspace{-10pt}
\crvspace{-8pt}
For the optimal transport, this leads to
\myvspace{-3pt}
\crvspace{-3pt}
\begin{equation}
\pi^\ast_{p,p_{\opn{draft}}}(i|j)=\begin{cases}\min(\frac{p(i)}{p_{\opn{draft}}(j)},1)&i=j\\(1-\frac{p(j)}{p_{\opn{draft}}(j)})_+\frac{(p(i)-p_{\opn{draft}}(i))_+}{\sum_{z\in\Sigma}(p_{\opn{draft}}(z)-p(z))_+}&i\neq j\end{cases}
~.
\end{equation}

\myvspace{-8pt}
\crvspace{-5pt}
The basic single-step, single-draft speculative decoding can be improved in two directions. Multi-step methods generate a draft sequence. Some improvements \citep{sun2024optimal,huaccelerated,sunblock} in this scenario are discussed in \Cref{sec:related_multi_step}. Our paper focuses on the multi-draft direction, where multiple draft tokens are generated at each step.

\myvspace{-7pt}
\crvspace{-7pt}
\subsection{Multi-Draft Speculative Decoding}\label{sec:back_mdsd}
\myvspace{-5pt}
\crvspace{-5pt}
For $i\in\Sigma$, define the incidence set $A_i:=\{\bar{i}\in \Sigma^n|\exists j\in[n], \overline{i}_j=i\}$.

Informally, the verification algorithm depends on two distributions $p$ and $p_{\opn{draft}}$, where $p_{\opn{draft}}$ is now a joint distribution of $n$ tokens. Common constructions of $p_{\opn{draft}}$ include:
\begin{itemize}[leftmargin=*,noitemsep=0mm,topsep=-2pt]
\item Sampling with replacement: Given a draft model with output distribution $q(\cdot)$, independently sample $n$ times. For $\bar{i}=(\bar{i}_1,\dots,\bar{i}_n)\in\Sigma^n$, we have $p_{\opn{draft}}(\bar{i})=\prod_{j=1}^n q(\bar{i}_j)$.
\item Sampling without replacement:
\begin{enumerate}[label=(\arabic*),leftmargin=*,noitemsep=0mm,topsep=0pt]
\item Successive sampling (SS): Draw one token at a time, removing each selected token and renormalizing $q$ before the next draw \citep{yu2012inclusion}. Thus $p_{\opn{draft}}(\bar{i})=\prod_{j=1}^n q^{\neg \bar{i}_1,\dots,\bar{i}_{j-1}}(\bar{i}_j)$, where
\begin{equation}\label{eq:successive}
q^{\neg \bar{i}_1,\dots,\bar{i}_{j-1}}(x)=\begin{cases}\frac{q(x)}{1-\sum_{z\in\{\bar{i}_1,\dots,\bar{i}_{j-1}\}}q(z)}&x\notin\{\bar{i}_1,\dots,\bar{i}_{j-1}\}\\0&x\in\{\bar{i}_1,\dots,\bar{i}_{j-1}\}\end{cases}
~.
\end{equation}
\item {Conditional Poisson sampling (CPS), also called rejective sampling \citep{hajek1964rejective,yu2012inclusion}: Draw $n$ tokens independently from $q$ and restart the entire batch whenever a duplicate occurs. The ordered draft tuple $\bar{i}=(\bar{i}_1,\dots,\bar{i}_n)$ has distribution
\begin{equation}\label{eq:cps_draft}
p_{\opn{draft}}(\bar{i})=
\begin{cases}
\displaystyle\frac{\prod_{j=1}^n q(\bar{i}_j)}{n!\,W_{n,\Sigma}} & \bar{i}_1,\dots,\bar{i}_n\text{ are distinct},\\[4pt]
0 & \text{otherwise}.
\end{cases}
\end{equation}
The corresponding set $S=\{\bar{i}_1,\dots,\bar{i}_n\}$ has the conditional Bernoulli distribution \citep[Eq.~(3)]{chen1997statistical}:
\begin{equation}\label{eq:cb_law}
\Pr(S)=\frac{\prod_{i\in S}q(i)}{W_{n,\Sigma}},\qquad |S|=n,
\qquad W_{k,H}:=\sum_{\substack{A\subseteq H\\|A|=k}}\prod_{i\in A}q(i),\quad W_{0,H}=1.
\end{equation}
}
\end{enumerate}
Both SS and CPS sample without replacement and require $1\leq n\leq|\operatorname{supp}(q)|$. They generally define different draft laws. The name \emph{without replacement} alone does not specify which one is used.
\item Product of different draft distributions: $p_{\opn{draft}}(\bar{i})=\prod_{j=1}^n q_j(\bar{i}_j)$.
\end{itemize}

\myvspace{-4pt}
Given multiple draft tokens $\bar{i}=(\bar{i}_1,\dots,\bar{i}_n)\sim p_{\opn{draft}}$, the goal is to output $i\sim p$ such that the objective $\max P(\exists j\in[n], i=\bar{i}_j)$ or equivalently $\max P(\bar{i} \in A_i)$ is achieved, that is to maximize the probability of random variable $i$ to be the same as one of random variable in $(\bar{i}_1,\dots,\bar{i}_n)$.

More formally, given $p\in \Delta_\Sigma$ and $p_{\opn{draft}}\in \Delta_{\Sigma^n}$ representing a probability distribution over the space $\Sigma$ and a probability distribution over the space $\Sigma^n$, respectively, we seek a joint distribution $\pi\in\Pi(p,p_{\opn{draft}})$ such that the marginal distributions are $p$ and $p_{\opn{draft}}$, respectively, and the objective $\max \sum_{i\in\Sigma} \sum_{\bar{i}\in A_i}\pi(i,\bar{i})$ is maximized. The optimal transport is denoted as $\pi^\ast_{p,p_{\opn{draft}}}\in\Pi(p,p_{\opn{draft}})$, and the optimal objective function value is $\alpha^\ast(p,p_{\opn{draft}})=\sum_{i\in\Sigma} \sum_{\bar{i}\in A_i}\pi^\ast_{p,p_{\opn{draft}}}(i,\bar{i})$.

The problem can be formulated as an LP by representing the joint distribution as a tensor:
\myvspace{-1pt}
\crvspace{-5pt}
\begin{equation}\label{eq:lp}
\begin{aligned}
\max_{C \in \mathbb{R}^{\Sigma \times \Sigma^n}} & \sum_{i \in \Sigma} \sum_{\bar{i} \in A_i} C_{i,\bar{i}} \\
\text{s.t.~} & \sum_{\bar{i} \in \Sigma^n} C_{i,\bar{i}} = p(i) \quad \forall i \in \Sigma, \quad \sum_{i \in \Sigma} C_{i,\bar{i}} = p_{\text{draft}}(\bar{i}) \quad \forall \bar{i} \in \Sigma^n, \\
& C_{i,\bar{i}} \geq 0 \quad \forall i \in \Sigma, \bar{i} \in \Sigma^n
.
\end{aligned}
\end{equation}
\par\myvspace{-10pt}
\crvspace{-10pt}
The difficulty lies in the exponential number of variables and constraints.

Several approximation have been proposed for the multi-draft speculative decoding problem, including Recursive Rejection Sampling (RRS) \citep{yang2024multi,jeon2024recursive} and K-SEQ \citep{sun2024spectr}. RRS recursively verifies the draft tokens, while K-SEQ improves the acceptance rate for sampling with replacement. Due to space constraints, we move the details of these methods (\Cref{se:approxverify}), other related work (\Cref{se:related_work}) and all proofs (\Cref{se:proofs}) to the appendix. \myvspace{-8pt}
\crvspace{-8pt}
\section{Optimal Acceptance Rate as Subset Selection Problem}\label{se:subset_selection}
\myvspace{-8pt}
\crvspace{-8pt}
We show that the optimal acceptance rate can be expressed as a subset selection problem:
\myvspace{-2pt}
\crvspace{-2pt}
\begin{equation}\label{eq:subset}
\alpha^\ast(p,p_{\opn{draft}})=
1+\min_{H\subset\Sigma} 
\biggl(
\sum_{i\in H} p(i)- \sum_{\bar{i}\in H^n} p_{\opn{draft}}(\bar{i})
\biggl)
~.
\end{equation}

\myvspace{-9pt}
\crvspace{-18pt}
\myvspace{-9pt}
\crvspace{-9pt}
\subsection{Dual Problem}
\myvspace{-4pt}
\crvspace{-4pt}
We start from the linear programming formulation \eqref{eq:lp} and derive an equivalent formulation:
\myvspace{-1pt}
\crvspace{-4pt}
\begin{equation}\label{eq:lp_prime}
\begin{aligned}
\max_{S \in \mathbb{R}^{\Sigma \times \Sigma^n}} & \sum_{i \in \Sigma} \sum_{\bar{i} \in \Sigma^n} S_{i,\bar{i}} \\
\text{s.t.~} & \sum_{\bar{i} \in \Sigma^n} S_{i,\bar{i}} \leq p(i) \quad\forall i \in \Sigma, \quad \sum_{i \in \Sigma} S_{i,\bar{i}} \leq p_{\text{draft}}(\bar{i}) \quad\forall \bar{i} \in \Sigma^n, \\
& S_{i,\bar{i}} \geq 0 \quad\forall i \in \Sigma, \bar{i} \in \Sigma^n, \quad S_{i,\bar{i}} = 0 \quad\forall i \in \Sigma, \bar{i} \notin A_i.
\end{aligned}
\end{equation}

\begin{lemma}\label{le:eq_lp_lp_prime}
    The two formulations \eqref{eq:lp} and \eqref{eq:lp_prime} are equivalent.
\end{lemma}
\myvspace{-8pt}
\crvspace{-8pt}
This equivalent formulation transforms the transportation problem \citep{hitchcock1941distribution} into a $b$-matching problem, whose dual is a $w$-vertex cover problem \citep{schrijver2003combinatorial} (with a detailed derivation in \Cref{se:dual}):
\myvspace{-8pt}
\crvspace{-3pt}
\begin{equation}\label{eq:dual}
\begin{aligned}
\min_{y \in \mathbb{R}^{\Sigma}, z \in \mathbb{R}^{\Sigma^n}} &
\sum_{i \in \Sigma} y_i p(i) + \sum_{\bar{i} \in \Sigma^n} z_{\bar{i}} p_{\text{draft}}(\bar{i}) 
\\
\text{s.t.~} & y_i + z_{\bar{i}} \geq 1 \quad \forall i \in \Sigma, \bar{i} \in A_i,\quad y_i \geq 0 \quad \forall i \in \Sigma, \quad z_{\bar{i}} \geq 0 \quad \forall \bar{i} \in \Sigma^n.
\end{aligned}
\end{equation}

\myvspace{-15pt}
\crvspace{-15pt}
\subsection{Total Unimodularity}
\myvspace{-4pt}
\crvspace{-4pt}
The coefficient matrix of the constraints in \eqref{eq:dual} is totally unimodular (TUM). The first set of constraints forms an incidence matrix of a bipartite graph, where one side of the nodes corresponds to $\Sigma$ and the other side corresponds to $\Sigma^n$. There is an edge between $i$ and $\bar{i}$ if and only if $\bar{i}\in A_i$. Therefore, it is a totally unimodular matrix \citep{biggs1993algebraic}. The second and third sets of constraints have coefficient matrices that are identity matrices, with $|\Sigma|+|\Sigma|^n$ variables and $|\Sigma|+|\Sigma|^n$ constraints. The concatenation of a TUM matrix and an identity matrix is also TUM \citep{commoner1973sufficient}.

Since the right-hand side of the constraints are integers, the dual problem \eqref{eq:dual} always has an integer optimal solution \citep{hoffman2010integral}.

\myvspace{-5pt}
\crvspace{-5pt}
\subsection{Subset Selection Formulation}
\myvspace{-5pt}
\crvspace{-3pt}
By restricting the variables in \eqref{eq:dual} to integers, we obtain:
\myvspace{-3pt}
\crvspace{-2pt}
\begin{equation}
\begin{aligned}
\min_{y\in\mathbb{Z}^{\Sigma},z\in\mathbb{Z}^{\Sigma^n}}  &
\sum_{i \in \Sigma} y_i p(i) + \sum_{\bar{i} \in \Sigma^n} z_{\bar{i}} p_{\text{draft}}(\bar{i}) 
\\
\text{s.t.~} & y_i + z_{\bar{i}} \geq 1 \quad \forall i \in \Sigma, \bar{i} \in A_i,\quad y_i \geq 0 \quad \forall i \in \Sigma, \quad z_{\bar{i}} \geq 0 \quad \forall \bar{i} \in \Sigma^n
.
\end{aligned}
\end{equation}
\par\myvspace{-5pt}
\crvspace{-3pt}
In the optimal solution, $y_i$ and $z_{\bar{i}}$ will not exceed 1, so they can only take values 0 or 1. Therefore, the problem can be further simplified as:
\myvspace{-3pt}
\crvspace{-1pt}
\begin{equation}
\begin{aligned}
\min_{y\in\{0,1\}^{\Sigma}}\min_{z\in\{0,1\}^{\Sigma^n}} & \sum_{i\in\Sigma} y_i p(i)+ \sum_{\bar{i}\in\Sigma^n} z_{\bar{i}} p_{\opn{draft}}(\bar{i})
\\
\text{s.t.~} & y_i+z_{\bar{i}}\geq1 \quad \forall i\in\Sigma,\bar{i}\in A_i
~.
\end{aligned}
\end{equation}
\par\myvspace{-5pt}
\crvspace{-3pt}
Define $H=\{i\in\Sigma|y_i=1\}$. The problem becomes:
\myvspace{-3pt}
\crvspace{-1pt}
\begin{equation}
\begin{aligned}
\min_{H\subset\Sigma}\min_{z\in\{0,1\}^{\Sigma^n}} & \sum_{i\in H} p(i)+ \sum_{\bar{i}\in\Sigma^n} z_{\bar{i}} p_{\opn{draft}}(\bar{i})
\\
\text{s.t.~} & z_{\bar{i}}\geq1  \quad \forall i\in\Sigma\setminus H,\bar{i}\in A_i
~.
\end{aligned}
\end{equation}
\par\myvspace{-5pt}
\crvspace{-3pt}
The optimal solution for $z$ is
\myvspace{-5pt}
\crvspace{-2pt}
\begin{equation}
z^\ast(H)_{\bar{i}}=\begin{cases}1&\bar{i}\in\bigcup_{x\in \Sigma\setminus H} A_x\\0&\bar{i}\notin\bigcup_{x\in \Sigma\setminus H} A_x\end{cases}
~.
\end{equation}
\par\myvspace{-5pt}
\crvspace{-2pt}
Substituting this solution, we obtain the subset selection formulation:
\myvspace{-3pt}
\begin{equation}
\min_{H\subset\Sigma}
\sum_{i\in H} p(i)+ \sum_{\bar{i}\in\bigcup_{x\in \Sigma\setminus H} A_x} p_{\opn{draft}}(\bar{i})
~.
\end{equation}
\par\myvspace{-5pt}
\crvspace{-2pt}
Finally, note that
\myvspace{-5pt}
\crvspace{-2pt}
\begin{equation}
\sum_{\bar{i}\in\bigcup_{x\in \Sigma\setminus H} A_x} p_{\opn{draft}}(\bar{i})+\sum_{\bar{i}\in H^n} p_{\opn{draft}}(\bar{i})=1
~.
\end{equation}
\par\myvspace{-5pt}
\crvspace{-2pt}
This completes the derivation of the subset selection formulation \eqref{eq:subset}. \crvspace{-5pt}
\section{Computing Optimal Acceptance Rate in Special Cases}\label{se:special_cases}
\crvspace{-5pt}
In this section, we discuss how to efficiently compute the optimal acceptance rate for certain special cases of the draft distribution $p_{\opn{draft}}$. For any set function $f$, we define the marginal value of an element $x$ with respect to a set $H$ as $f(x|H)=f(H\cup \{x\})-f(H)$. 
We also define the following shorthand notations: $P(H)=\sum_{i\in H} p(i)$, $Q(H)=\sum_{\bar{i}\in H^n} p_{\opn{draft}}(\bar{i})$, $f(H)=P(H)-Q(H)$.

The optimal acceptance rate can be expressed as
$\alpha^\ast(p,p_{\opn{draft}})=1+\min_{H\subset\Sigma} f(H)$.

\myvspace{-5pt}
\crvspace{-3pt}
\subsection{\texorpdfstring{$q$}{q}-Convex Functions}
\myvspace{-5pt}
\crvspace{-2pt}
\begin{definition}[$q$-Convex Function]
A set function $Q:2^\Sigma\to\mathbb{R}$ is called a $q$-convex function if there exists a function $q:\Sigma\to\mathbb{R}_{\geq0}$ such that for all $H\subset \Sigma$ and $x,y\in \Sigma\setminus H$ with $x\neq y$, we have
\begin{equation}
\frac{Q(x|H)}{q(x)}\leq\frac{Q(y|H\cup\{x\})}{q(y)}
~.
\end{equation}
\end{definition}
\par\crvspace{-3pt}
Intuitively, if we order the elements of $\Sigma$ arbitrarily and construct a sequence of sets $H_i$ by adding elements one by one, then the curve of $Q(H_i)$ against the sum of $q$ values is always convex.

\begin{theorem}\label{th:q_convex_wr}
For sampling with replacement, the function $Q$ is a $q$-convex function.
\end{theorem}

\begin{theorem}[Successive sampling]\label{th:q_convex_wo}
For successive sampling as defined in \eqref{eq:successive}, with $1\leq n\leq|\operatorname{supp}(q)|$, the function $Q$ is $q$-convex.
\end{theorem}
\begin{theorem}[Conditional Poisson sampling]\label{th:q_convex_cb}
For conditional Poisson sampling as defined in \eqref{eq:cb_law}, with $1\leq n\leq|\operatorname{supp}(q)|$, the function $Q$ is $q$-convex.
\end{theorem}

\begin{theorem}\label{th:q_convex_supermodular}
All $q$-convex functions are supermodular functions.
\end{theorem}

For sampling with replacement, SS, and CPS, computing $\alpha^\ast(p,p_{\opn{draft}})$ is an unconstrained submodular minimization problem, solvable with polynomially many value-oracle queries \citep{iwata2008submodular}. The next section reduces the search to sorted prefixes. The total cost also depends on evaluating the corresponding $Q(H)$.

\subsection{Efficient Computation}\label{se:efficient}
\begin{theorem}\label{th:efficient_computation}
Suppose that $Q$ is a $q$-convex function, $Q$ is monotone increasing, and $p(x)>0$ for all $x\in\Sigma$. For all $H\subset \Sigma$ and $x,y\in \Sigma\setminus H$ with $x\neq y$, if $\frac{q(x)}{p(x)}\leq\frac{q(y)}{p(y)}$ and $f(x|H)\leq0$, then $f(y|H\cup\{x\})\leq0$.
\end{theorem}

The above theorem requires $p(x)>0$. When $p(x)=0$, there exists an optimal set $H^\ast$ for \eqref{eq:subset} that contains $x$ because $Q$ is monotone increasing.

\subsubsection{Algorithm}
Inspired by \Cref{th:efficient_computation}, we can compute the optimal acceptance rate efficiently as follows:
\begin{enumerate}[leftmargin=*,noitemsep=0mm,topsep=-2pt]
\item Find an ordering $\sigma$ of $\Sigma$ such that $\frac{q(\sigma_1)}{p(\sigma_1)}\geq\dots\geq\frac{q(\sigma_{|\Sigma|})}{p(\sigma_{|\Sigma|})}$.
\item Construct a sequence of sets $H_i = \{\sigma_1,\dots,\sigma_i\}$.
\item Compute $\alpha^\ast(p,p_{\opn{draft}})=
1+\min_{i} f(H_i)$.
\end{enumerate}

Intuitively, we sort the elements by the ratio of $q$ and $p$ in non-increasing order and then perform a linear search.

\subsubsection{Complexity of Computing \texorpdfstring{$Q$}{Q} and \texorpdfstring{$\alpha^\ast$}{alpha*}}
For sampling with replacement, $Q$ has a simple expression $Q(H)=(\sum_{x\in H} q(x))^n$. The time complexity for computing $\alpha^\ast(p,p_{\opn{draft}})$ is $O(|\Sigma|\log |\Sigma|)$ for the sorting step, plus $O(|\Sigma|)$ for the linear scan.

For conditional Poisson sampling, $Q(H)=W_{n,H}/W_{n,\Sigma}$, where $W_{n,H}=\operatorname{Coeff}_{t^n}\prod_{i\in H}(1+q(i)t)$. The recurrence $W_{k,H\cup\{x\}}=W_{k,H}+q(x)W_{k-1,H}$, initialized by $W_{0,H}=1$ and $W_{k,\varnothing}=0$ for $k>0$, evaluates all prefixes in $O(n|\Sigma|)$ arithmetic operations. Including sorting, the cost is $O(|\Sigma|\log|\Sigma|+n|\Sigma|)$.

For successive sampling, the same prefix reduction holds by \Cref{th:q_convex_wo}, but $Q(H)$ must be evaluated under successive sampling.
 \crvspace{-5pt}
\section{A Greedy Approach For Selecting Draft Tokens}\label{se:greedy_approach}
\crvspace{-5pt}
In this section, we propose a novel method for constructing the draft distribution $p_{\opn{draft}}$ and a corresponding verification algorithm that achieves the optimal acceptance rate for this distribution.

\crvspace{-5pt}
\subsection{Draft Construction}
\crvspace{-3pt}
Given a draft model output distribution $q\in\Delta_\Sigma$, we construct the draft tokens $\bar{i}=(\bar{i}_1,\dots,\bar{i}_n)$ as follows:
\begin{itemize}[leftmargin=*,noitemsep=0mm,topsep=-2pt]
\item The first $n-1$ tokens are deterministically set to be the top $n-1$ tokens according to the probability in $q$, i.e., $\bar{i}_1,\dots,\bar{i}_{n-1}=\opn{Top}_{n-1}(q)$, such that $q(\bar{i}_1)\geq\dots\geq q(\bar{i}_{n-1})$ and $\max_{i\in\Sigma\setminus\{\bar{i}_1,\dots,\bar{i}_{n-1}\}} q(i)\leq q(\bar{i}_{n-1})$.
\item Only the last token $\bar{i}_{n}$ is randomly sampled from $q$ without replacement (i.e., it is different from the previous $n-1$ tokens): $\bar{i}_{n}\sim q^{\neg \opn{Top}_{n-1}(q)}(\cdot)=\frac{q(\cdot)}{1-\sum_{j=1}^{n-1}q(\bar{i}_j)}$.
\end{itemize}

The resulting draft distribution is
\begin{equation}
p_{\opn{draft}}(\bar{i})=\begin{cases}q^{\neg \opn{Top}_{n-1}(q)}(\bar{i}_n)&\bar{i}_1,\dots,\bar{i}_{n-1}=\opn{Top}_{n-1}(q)\\0&\bar{i}_1,\dots,\bar{i}_{n-1}\neq\opn{Top}_{n-1}(q)\end{cases}
~.
\end{equation}

\subsection{Verification Algorithm}
The corresponding optimal transport problem for this draft distribution is simple because only one draft token is random. We can design a verification algorithm that strictly achieves the optimal acceptance rate for this draft distribution (, with unfolded definition in \Cref{sec:add_ill}):
\begin{equation}
\pi^{\opn{Greedy}}_{p,p_{\opn{draft}}}(i|\bar{i})=\pi^\ast_{p,q^{\neg \opn{Top}_{n-1}(q)}}(i|\bar{i}_n)
~.
\end{equation}

\begin{theorem}\label{th:greedy_optimal}
The optimal acceptance rate for the greedy draft distribution is
\begin{equation}
\alpha^\ast(p,p_{\opn{draft}})=\alpha^{\opn{Greedy}}(p,p_{\opn{draft}})=
\sum_{i\in\opn{Top}_{n-1}(q)}p(i)
+\sum_{i\in\Sigma}\min(p(i),q^{\neg \opn{Top}_{n-1}(q)}(i))
~.
\end{equation}
\end{theorem}
\par\crvspace{-5pt}
Our subset selection formulation \eqref{eq:subset} provides a convenient way to prove the above theorem.
\subsection{Connection to SpecHub}\label{se:connection_to_spechub}
SpecHub \citep{anonymous2024spechub} is a recently proposed MDSD method that is only applicable to the case of $n=2$. The draft construction in SpecHub is as follows:
\begin{itemize}[leftmargin=*,noitemsep=0mm,topsep=-2pt]
\item First, sample the first draft token $\bar{i}_1$.
\item If $\bar{i}_1=\opn{Top}_1(q)$ is the token with the highest probability in $q$, then sample the second draft token $\bar{i}_2$ without replacement to ensure it is different from $\bar{i}_1$.
\item If $\bar{i}_1\neq\opn{Top}_1(q)$ is not the token with the highest probability in $q$, then deterministically set the second draft token to be the token with the highest probability, i.e., $\bar{i}_2=\opn{Top}_1(q)$.
\end{itemize}

The resulting draft distribution is:
\crvspace{-3pt}
\begin{equation}
p_{\opn{draft}}(\bar{i})=\begin{cases}
q(\bar{i}_1)&\bar{i}_2=\opn{Top}_1(q)\\
\frac{q(\bar{i}_1)}{1-q(\bar{i}_1)}q(\bar{i}_2)&\bar{i}_1=\opn{Top}_1(q)\\
0&\text{otherwise}\end{cases}
~.
\end{equation}
\par\crvspace{-3pt}
We note that the greedy method for $n=2$ is essentially equivalent to SpecHub because both methods ensure that at least one draft token is the token with the highest probability in $q$. However, the specific draft distributions are different, leading to a simpler verification algorithm for the greedy method. \section{Experiments}\label{se:experiments}

\begin{table}[ht]
\small
\centering
\setlength{\tabcolsep}{4pt}
\caption{
Acceptance rates of different MDSD methods across various models and tasks. $\Delta \alpha$ means the difference between a verification method and the displayed theoretical upper bound, with statistically significant differences indicated by directional arrows. For RRS (SS), this is a comparison with the CPS optimum, not an optimality gap.
}\label{tab:main}
\resizebox{1\linewidth}{!}{

\begin{tabular}{lllcccccc}
\toprule
\multirow{2}{*}{Model Pairs} & \multirow{2}{*}{Draft Sampling} & \multirow{2}{*}{Method} & \multicolumn{2}{c}{Alpaca} & \multicolumn{2}{c}{CNN-DailyMail} & \multicolumn{2}{c}{WMT'14}\\

\cmidrule(r){4 - 5}\cmidrule(r){6 - 7}\cmidrule(r){8 - 9}
&  &  & $\alpha$ & $\Delta \alpha$ & $\alpha$ & $\Delta \alpha$ & $\alpha$ & $\Delta \alpha$ \\
\midrule
\multirow{8}{*}{\makecell[l]{OPT-125M \\ OPT-6.7B}} & \multirow{4}{*}{\makecell[l]{With \\ Replacement}} & RRS & $85.4 \pm 0.1$ & $-1.5\downarrow$ & $77.3 \pm 0.1$ & $-3.3\downarrow$ & $70.6 \pm 0.1$ & $-1.5\downarrow$ \\
 &  & K-SEQ & $85.8 \pm 0.1$ & $-1.1\downarrow$ & $78.4 \pm 0.1$ & $-2.2\downarrow$ & $71.1 \pm 0.1$ & $-0.9\downarrow$ \\
 &  & $\alpha^{\opn{K-SEQ}}$ & $85.9 \pm 0.1$ & $-0.9\downarrow$ & $78.5 \pm 0.1$ & $-2.2\downarrow$ & $71.0 \pm 0.1$ & $-1.0\downarrow$ \\
 &  & $\alpha^\ast$ & $\bm{86.9} \pm 0.1$ & - & $\bm{80.7} \pm 0.1$ & - & $\bm{72.0} \pm 0.1$ & - \\
\cmidrule(r){2-9} & \multirow{2}{*}{\makecell[l]{Without \\ Replacement}} & RRS (SS) & $88.9 \pm 0.1$ & $-0.9\downarrow$ & $81.5 \pm 0.1$ & $-2.8\downarrow$ & $75.1 \pm 0.1$ & $-1.0\downarrow$ \\
 &  & $\alpha^\ast_{\mathrm{CPS}}$ & $\bm{89.9} \pm 0.1$ & - & $\bm{84.3} \pm 0.1$ & - & $\bm{76.0} \pm 0.1$ & - \\
\cmidrule(r){2-9} & \multirow{2}{*}{\makecell[l]{Greedy}} & Verify & $\bm{90.7} \pm 0.1$ & $0.0$ & $\bm{84.2} \pm 0.1$ & $-0.1$ & $\bm{77.0} \pm 0.1$ & $-0.0$ \\
 &  & $\alpha^\ast$ & $\bm{90.7} \pm 0.1$ & - & $\bm{84.3} \pm 0.1$ & - & $\bm{77.1} \pm 0.1$ & - \\
\midrule
\multirow{8}{*}{\makecell[l]{LLaMA-68M \\ LLaMA-7B}} & \multirow{4}{*}{\makecell[l]{With \\ Replacement}} & RRS & $71.6 \pm 0.1$ & $-1.5\downarrow$ & $65.3 \pm 0.1$ & $-2.2\downarrow$ & $59.8 \pm 0.1$ & $-1.0\downarrow$ \\
 &  & K-SEQ & $71.9 \pm 0.1$ & $-1.1\downarrow$ & $66.0 \pm 0.1$ & $-1.5\downarrow$ & $60.0 \pm 0.1$ & $-0.8\downarrow$ \\
 &  & $\alpha^{\opn{K-SEQ}}$ & $72.0 \pm 0.1$ & $-1.0\downarrow$ & $66.2 \pm 0.1$ & $-1.3\downarrow$ & $60.2 \pm 0.1$ & $-0.6\downarrow$ \\
 &  & $\alpha^\ast$ & $\bm{73.0} \pm 0.1$ & - & $\bm{67.5} \pm 0.1$ & - & $\bm{60.8} \pm 0.1$ & - \\
\cmidrule(r){2-9} & \multirow{2}{*}{\makecell[l]{Without \\ Replacement}} & RRS (SS) & $75.7 \pm 0.1$ & $-0.8\downarrow$ & $70.5 \pm 0.1$ & $-1.3\downarrow$ & $\bm{63.3} \pm 0.1$ & $-0.1$ \\
 &  & $\alpha^\ast_{\mathrm{CPS}}$ & $\bm{76.5} \pm 0.1$ & - & $\bm{71.8} \pm 0.1$ & - & $\bm{63.4} \pm 0.1$ & - \\
\cmidrule(r){2-9} & \multirow{2}{*}{\makecell[l]{Greedy}} & Verify & $\bm{78.4} \pm 0.1$ & $-0.1$ & $\bm{73.2} \pm 0.1$ & $0.1$ & $\bm{66.1} \pm 0.1$ & $0.0$ \\
 &  & $\alpha^\ast$ & $\bm{78.4} \pm 0.1$ & - & $\bm{73.1} \pm 0.1$ & - & $\bm{66.1} \pm 0.1$ & - \\
\midrule
\multirow{8}{*}{\makecell[l]{Eagle-0.24B \\ Vicuna-7B}} & \multirow{4}{*}{\makecell[l]{With \\ Replacement}} & RRS & $63.4 \pm 0.2$ & $-1.0\downarrow$ & $56.7 \pm 0.1$ & $-1.1\downarrow$ & $\bm{32.9} \pm 0.2$ & $-0.2$ \\
 &  & K-SEQ & $63.9 \pm 0.2$ & $-0.5\downarrow$ & $57.0 \pm 0.1$ & $-0.8\downarrow$ & $\bm{33.0} \pm 0.2$ & $-0.1$ \\
 &  & $\alpha^{\opn{K-SEQ}}$ & $63.7 \pm 0.1$ & $-0.7\downarrow$ & $57.1 \pm 0.1$ & $-0.7\downarrow$ & $\bm{32.9} \pm 0.2$ & $-0.1$ \\
 &  & $\alpha^\ast$ & $\bm{64.4} \pm 0.1$ & - & $\bm{57.8} \pm 0.1$ & - & $\bm{33.1} \pm 0.2$ & - \\
\cmidrule(r){2-9} & \multirow{2}{*}{\makecell[l]{Without \\ Replacement}} & RRS (SS) & $70.9 \pm 0.2$ & $-0.6\downarrow$ & $63.4 \pm 0.1$ & $-0.8\downarrow$ & $\bm{36.9} \pm 0.2$ & $0.4$ \\
 &  & $\alpha^\ast_{\mathrm{CPS}}$ & $\bm{71.5} \pm 0.1$ & - & $\bm{64.2} \pm 0.1$ & - & $\bm{36.4} \pm 0.2$ & - \\
\cmidrule(r){2-9} & \multirow{2}{*}{\makecell[l]{Greedy}} & Verify & $\bm{72.6} \pm 0.2$ & $-0.2$ & $\bm{65.7} \pm 0.1$ & $-0.1$ & $\bm{39.5} \pm 0.2$ & $0.1$ \\
 &  & $\alpha^\ast$ & $\bm{72.8} \pm 0.1$ & - & $\bm{65.8} \pm 0.1$ & - & $\bm{39.5} \pm 0.2$ & - \\
\midrule
\multirow{8}{*}{\makecell[l]{Eagle-0.26B \\ Qwen2-7B}} & \multirow{4}{*}{\makecell[l]{With \\ Replacement}} & RRS & $59.6 \pm 0.2$ & $-1.0\downarrow$ & $46.7 \pm 0.1$ & $-1.6\downarrow$ & $38.3 \pm 0.1$ & $-0.4\downarrow$ \\
 &  & K-SEQ & $59.9 \pm 0.2$ & $-0.8\downarrow$ & $47.2 \pm 0.1$ & $-1.1\downarrow$ & $\bm{38.3} \pm 0.1$ & $-0.4$ \\
 &  & $\alpha^{\opn{K-SEQ}}$ & $59.9 \pm 0.1$ & $-0.7\downarrow$ & $47.3 \pm 0.1$ & $-1.0\downarrow$ & $\bm{38.3} \pm 0.1$ & $-0.3$ \\
 &  & $\alpha^\ast$ & $\bm{60.7} \pm 0.1$ & - & $\bm{48.3} \pm 0.1$ & - & $\bm{38.7} \pm 0.1$ & - \\
\cmidrule(r){2-9} & \multirow{2}{*}{\makecell[l]{Without \\ Replacement}} & RRS (SS) & $68.3 \pm 0.2$ & $-1.1\downarrow$ & $52.4 \pm 0.1$ & $-1.7\downarrow$ & $\bm{43.9} \pm 0.1$ & $-0.1$ \\
 &  & $\alpha^\ast_{\mathrm{CPS}}$ & $\bm{69.4} \pm 0.1$ & - & $\bm{54.1} \pm 0.1$ & - & $\bm{44.0} \pm 0.1$ & - \\
\cmidrule(r){2-9} & \multirow{2}{*}{\makecell[l]{Greedy}} & Verify & $\bm{69.9} \pm 0.2$ & $-0.0$ & $\bm{54.0} \pm 0.1$ & $0.0$ & $\bm{45.5} \pm 0.1$ & $0.1$ \\
 &  & $\alpha^\ast$ & $\bm{70.0} \pm 0.1$ & - & $\bm{53.9} \pm 0.1$ & - & $\bm{45.4} \pm 0.1$ & - \\

\bottomrule
\end{tabular} }
\crvspace{-10pt}
\end{table} 
The goal of our experiments is to measure the acceptance rates of various MDSD methods on real text distributions and compare them with the theoretical upper bounds. In the previous sections, we analyzed the theoretical acceptance rate $\alpha^\ast(p,p_{\opn{draft}})$ for three different draft distributions: sampling with replacement, conditional Poisson sampling, and greedy approach (\Cref{se:greedy_approach}). We also discussed some existing verification methods (\Cref{se:approxverify}), such as RRS and K-SEQ, whose acceptance rates are expected to be lower than the theoretical upper bound for the same draft distribution. For K-SEQ, its average acceptance rate $\alpha^{\opn{K-SEQ}}$ can be derived theoretically (see \Cref{se:kseq} for details). Our efficient computation methods (\Cref{se:special_cases}) make it possible, for the first time, to obtain the theoretical upper bound of MDSD for vocabulary sizes of thousands.

To obtain realistic distributions $p$ and $p_{\opn{draft}}$, we select real-world datasets for various tasks, including Alpaca \citep{alpaca} for instruction-following, WMT'14 De-En \citep{bojar2014findings} for translation, and CNN-DailyMail \citep{DBLP:conf/nips/HermannKGEKSB15} for summarization. For each task, we use an LLM to generate responses on 1024 data samples, with a maximum length of 128 tokens. We then measure the logits of the target model and the draft model on these generated responses to construct $p$ and $p_{\opn{draft}}$.

We evaluated different approaches based on four publicly available large language models, including 1) LLaMA~\citep{touvron2023llama}, 2) Vicuna~\citep{vicuna2023}, the instruction fine-tuned version of LLaMA models, 3) OPT~\citep{zhang2022opt}, and 4) Qwen2~\citep{yang2024qwen2}. Specifically, for the LLaMA family, we select LLaMA-7B as the target model and LLaMA-68M as the draft model, which is consistent with previous work~\citep{miao2024specinfer}. For the OPT family, we select OPT-6.7B as the target model and OPT-125M as the draft model. Moreover, for the Vicuna family and the Qwen family, we select Vicuna-7B-v1.3 and Qwen2-7B-Instruct as target models, and we use paired draft models provided by EAGEL \citep{li2024eagle}, with 0.24B parameters and 0.26B parameters, respectively.

Unless otherwise specified, we use a default generation temperature of $0.7$ and a draft token number of $3$. The total computational cost is less than 50 GPU hours on RTXA6000.

\crvspace{-10pt}
\subsection{Main Experiment}
\crvspace{-5pt}
In the main experiment, we compare the acceptance rates of different MDSD methods across various LLMs and tasks. The results are shown in \Cref{tab:main}. We observe that, for sampling with replacement, the existing verify methods, RRS and K-SEQ, still have gaps compared to the theoretical acceptance rate upper bound. Sampling without replacement achieves higher acceptance rates than sampling with replacement, using CPS for the theoretical upper bound and SS for RRS. We can attribute this to the fact that sampling with replacement may lead to duplicate draft tokens, which are less helpful for acceleration. The greedy method obtains the highest acceptance rate, but this is not always the case, as we will see in the ablation study below that the greedy method performs worse when the temperature is $1$.

\begin{figure}[ht]
\pgfplotsset{
    GCSpS_theory_style/.style={
        color=red, thick, densely dotted, mark=triangle*, mark options={solid, scale=0.5}
    },
    wor_optimal_theory_style/.style={
        color=green!70!black, thick, dashed, mark=triangle*, mark options={solid, scale=0.5}
    },
    wr_optimal_theory_style/.style={
        color=magenta, thick, dash dot, mark=triangle*, mark options={solid, scale=0.5}
    },
    K_Seq_theory_style/.style={
        color=blue, thick, dotted, mark=triangle*, mark options={solid, scale=0.5}
    },
    GCSpS_verify_style/.style={
        color=orange, thick, solid, mark=square*, mark options={solid, scale=0.5}
    },
    wor_recursive_verify_style/.style={
        color=purple, thick, thick, mark=pentagon*, mark options={solid, scale=0.5}
    },
    wr_recursive_verify_style/.style={
        color=black, thick, solid, mark=x, mark options={solid, scale=0.8}
    },
    K_Seq_verify_style/.style={
        color=cyan, thick, solid, mark=diamond*, mark options={solid, scale=0.5}
    }
}
\centering
\begin{tikzpicture}
    \begin{axis}[
        hide axis, xmin=0, xmax=1, ymin=0, ymax=1, legend columns=4, legend style={/tikz/every even column/.append style={column sep=0.5em}}, width=10cm, height=2cm, legend image post style={xscale=0.5}, ]
   
    \addplot[GCSpS_theory_style] coordinates {(0,0)};
    \addlegendentry{\scalebox{0.8}{Greedy Optimal}}
    
    \addplot[wor_optimal_theory_style] coordinates {(0,0)};
    \addlegendentry{\scalebox{0.8}{CPS Optimal}}
    
    \addplot[wr_optimal_theory_style] coordinates {(0,0)};
    \addlegendentry{\scalebox{0.8}{With Replacement Optimal}}

    \addplot[K_Seq_theory_style] coordinates {(0,0)};
    \addlegendentry{\scalebox{0.8}{K-SEQ Theory}}

    \addplot[GCSpS_verify_style] coordinates {(0,0)};
    \addlegendentry{\scalebox{0.8}{Greedy}}
    
    \addplot[wor_recursive_verify_style] coordinates {(0,0)};
    \addlegendentry{\scalebox{0.8}{Successive RRS}}
     
    \addplot[wr_recursive_verify_style] coordinates {(0,0)};
    \addlegendentry{\scalebox{0.8}{With Replacement RRS}}
    
    \addplot[K_Seq_verify_style] coordinates {(0,0)};
    \addlegendentry{\scalebox{0.8}{K-SEQ}}
    
    \end{axis}
\end{tikzpicture}
 
\begin{subfigure}[b]{0.36\textwidth} \begin{tikzpicture}
\begin{axis}[
    width=\textwidth,
    height=0.8\textwidth,
    xlabel={Temperatures},
    xlabel style={font=\scriptsize, yshift=0.5em}, ylabel style={font=\scriptsize, yshift=-1.5em}, ylabel={$\alpha$},
    xtick={0.1, 0.3, 0.5, 0.7, 0.9},
    xmin=-0.1, xmax=1.1,
xticklabel style={font=\scriptsize},
    yticklabel style={font=\scriptsize},
    grid=major,
    legend style={font=\scriptsize, at={(1.05,1)}, anchor=north west},
]
\addplot[name path=K_Seq_verify_upper, draw=none] coordinates {
(0.0, 0.660) (0.1, 0.667) (0.3, 0.668) (0.5, 0.637) (0.7, 0.601) (0.72, 0.599) (0.74, 0.593) (0.76, 0.591) (0.78, 0.590) (0.8, 0.586) (0.82, 0.578) (0.84, 0.576) (0.86, 0.572) (0.88, 0.566) (0.9, 0.565) (0.92, 0.566) (0.94, 0.573) (0.96, 0.578) (0.98, 0.579) (1, 0.587) 
};
\addplot[name path=K_Seq_verify_lower, draw=none] coordinates {
(0.0, 0.658) (0.1, 0.665) (0.3, 0.666) (0.5, 0.634) (0.7, 0.599) (0.72, 0.596) (0.74, 0.591) (0.76, 0.588) (0.78, 0.587) (0.8, 0.583) (0.82, 0.575) (0.84, 0.573) (0.86, 0.569) (0.88, 0.563) (0.9, 0.563) (0.92, 0.563) (0.94, 0.570) (0.96, 0.575) (0.98, 0.577) (1, 0.584) 
};
\addplot[K_Seq_verify_style, opacity=0.2] fill between[of=K_Seq_verify_upper and K_Seq_verify_lower];

\addplot[K_Seq_verify_style, mark indices={1,2,3,4,5,13,20}] coordinates {
(0.0, 0.659) (0.1, 0.666) (0.3, 0.667) (0.5, 0.636) (0.7, 0.600) (0.72, 0.598) (0.74, 0.592) (0.76, 0.589) (0.78, 0.588) (0.8, 0.584) (0.82, 0.577) (0.84, 0.574) (0.86, 0.570) (0.88, 0.564) (0.9, 0.564) (0.92, 0.564) (0.94, 0.572) (0.96, 0.576) (0.98, 0.578) (1, 0.586) 
};

\addplot[name path=wor_recursive_verify_upper, draw=none] coordinates {
(0.0, 0.765) (0.1, 0.762) (0.3, 0.740) (0.5, 0.688) (0.7, 0.634) (0.72, 0.629) (0.74, 0.620) (0.76, 0.615) (0.78, 0.613) (0.8, 0.605) (0.82, 0.597) (0.84, 0.592) (0.86, 0.585) (0.88, 0.582) (0.9, 0.578) (0.92, 0.576) (0.94, 0.579) (0.96, 0.582) (0.98, 0.583) (1, 0.587) 
};
\addplot[name path=wor_recursive_verify_lower, draw=none] coordinates {
(0.0, 0.763) (0.1, 0.760) (0.3, 0.737) (0.5, 0.686) (0.7, 0.632) (0.72, 0.626) (0.74, 0.617) (0.76, 0.612) (0.78, 0.610) (0.8, 0.602) (0.82, 0.594) (0.84, 0.589) (0.86, 0.582) (0.88, 0.579) (0.9, 0.576) (0.92, 0.573) (0.94, 0.576) (0.96, 0.579) (0.98, 0.580) (1, 0.584) 
};
\addplot[wor_recursive_verify_style, opacity=0.2] fill between[of=wor_recursive_verify_upper and wor_recursive_verify_lower];

\addplot[wor_recursive_verify_style, mark indices={1,2,3,4,5,13,20}] coordinates {
(0.0, 0.764) (0.1, 0.761) (0.3, 0.738) (0.5, 0.687) (0.7, 0.633) (0.72, 0.627) (0.74, 0.619) (0.76, 0.614) (0.78, 0.611) (0.8, 0.603) (0.82, 0.595) (0.84, 0.590) (0.86, 0.584) (0.88, 0.581) (0.9, 0.577) (0.92, 0.574) (0.94, 0.577) (0.96, 0.580) (0.98, 0.582) (1, 0.585) 
};

\addplot[name path=GCSpS_verify_upper, draw=none] coordinates {
(0.0, 0.765) (0.1, 0.763) (0.3, 0.742) (0.5, 0.700) (0.7, 0.662) (0.72, 0.656) (0.74, 0.651) (0.76, 0.644) (0.78, 0.642) (0.8, 0.637) (0.82, 0.629) (0.84, 0.624) (0.86, 0.615) (0.88, 0.610) (0.9, 0.602) (0.92, 0.597) (0.94, 0.596) (0.96, 0.595) (0.98, 0.593) (1, 0.592) 
};
\addplot[name path=GCSpS_verify_lower, draw=none] coordinates {
(0.0, 0.763) (0.1, 0.760) (0.3, 0.740) (0.5, 0.698) (0.7, 0.659) (0.72, 0.654) (0.74, 0.648) (0.76, 0.641) (0.78, 0.640) (0.8, 0.634) (0.82, 0.627) (0.84, 0.621) (0.86, 0.613) (0.88, 0.607) (0.9, 0.599) (0.92, 0.594) (0.94, 0.593) (0.96, 0.592) (0.98, 0.590) (1, 0.589) 
};
\addplot[GCSpS_verify_style, opacity=0.2] fill between[of=GCSpS_verify_upper and GCSpS_verify_lower];

\addplot[GCSpS_verify_style, mark indices={1,2,3,4,5,13,20}] coordinates {
(0.0, 0.764) (0.1, 0.761) (0.3, 0.741) (0.5, 0.699) (0.7, 0.661) (0.72, 0.655) (0.74, 0.649) (0.76, 0.643) (0.78, 0.641) (0.8, 0.636) (0.82, 0.628) (0.84, 0.623) (0.86, 0.614) (0.88, 0.609) (0.9, 0.600) (0.92, 0.596) (0.94, 0.595) (0.96, 0.594) (0.98, 0.591) (1, 0.590) 
};

\addplot[name path=wr_recursive_verify_upper, draw=none] coordinates {
(0.0, 0.660) (0.1, 0.667) (0.3, 0.669) (0.5, 0.636) (0.7, 0.599) (0.72, 0.596) (0.74, 0.588) (0.76, 0.586) (0.78, 0.585) (0.8, 0.580) (0.82, 0.573) (0.84, 0.568) (0.86, 0.565) (0.88, 0.561) (0.9, 0.556) (0.92, 0.557) (0.94, 0.563) (0.96, 0.567) (0.98, 0.569) (1, 0.573) 
};
\addplot[name path=wr_recursive_verify_lower, draw=none] coordinates {
(0.0, 0.658) (0.1, 0.665) (0.3, 0.666) (0.5, 0.633) (0.7, 0.596) (0.72, 0.593) (0.74, 0.585) (0.76, 0.583) (0.78, 0.582) (0.8, 0.577) (0.82, 0.570) (0.84, 0.565) (0.86, 0.562) (0.88, 0.558) (0.9, 0.553) (0.92, 0.554) (0.94, 0.560) (0.96, 0.564) (0.98, 0.566) (1, 0.571) 
};
\addplot[wr_recursive_verify_style, opacity=0.2] fill between[of=wr_recursive_verify_upper and wr_recursive_verify_lower];

\addplot[wr_recursive_verify_style, mark indices={1,2,3,4,5,13,20}] coordinates {
(0.0, 0.659) (0.1, 0.666) (0.3, 0.667) (0.5, 0.634) (0.7, 0.598) (0.72, 0.595) (0.74, 0.587) (0.76, 0.584) (0.78, 0.584) (0.8, 0.578) (0.82, 0.571) (0.84, 0.567) (0.86, 0.564) (0.88, 0.559) (0.9, 0.555) (0.92, 0.555) (0.94, 0.561) (0.96, 0.565) (0.98, 0.567) (1, 0.572) 
};

\addplot[name path=K_Seq_theory_upper, draw=none] coordinates {
(0.0, 0.660) (0.1, 0.667) (0.3, 0.668) (0.5, 0.636) (0.7, 0.603) (0.72, 0.599) (0.74, 0.593) (0.76, 0.590) (0.78, 0.589) (0.8, 0.585) (0.82, 0.578) (0.84, 0.576) (0.86, 0.570) (0.88, 0.569) (0.9, 0.565) (0.92, 0.567) (0.94, 0.571) (0.96, 0.575) (0.98, 0.578) (1, 0.585) 
};
\addplot[name path=K_Seq_theory_lower, draw=none] coordinates {
(0.0, 0.658) (0.1, 0.665) (0.3, 0.666) (0.5, 0.634) (0.7, 0.601) (0.72, 0.597) (0.74, 0.591) (0.76, 0.587) (0.78, 0.587) (0.8, 0.583) (0.82, 0.575) (0.84, 0.574) (0.86, 0.568) (0.88, 0.567) (0.9, 0.563) (0.92, 0.565) (0.94, 0.569) (0.96, 0.573) (0.98, 0.577) (1, 0.583) 
};
\addplot[K_Seq_theory_style, opacity=0.2] fill between[of=K_Seq_theory_upper and K_Seq_theory_lower];

\addplot[K_Seq_theory_style, mark indices={1,2,3,4,5,13,20}] coordinates {
(0.0, 0.659) (0.1, 0.666) (0.3, 0.667) (0.5, 0.635) (0.7, 0.602) (0.72, 0.598) (0.74, 0.592) (0.76, 0.589) (0.78, 0.588) (0.8, 0.584) (0.82, 0.577) (0.84, 0.575) (0.86, 0.569) (0.88, 0.568) (0.9, 0.564) (0.92, 0.566) (0.94, 0.570) (0.96, 0.574) (0.98, 0.578) (1, 0.584) 
};

\addplot[name path=GCSpS_theory_upper, draw=none] coordinates {
(0.0, 0.765) (0.1, 0.762) (0.3, 0.743) (0.5, 0.701) (0.7, 0.662) (0.72, 0.657) (0.74, 0.651) (0.76, 0.646) (0.78, 0.644) (0.8, 0.639) (0.82, 0.629) (0.84, 0.625) (0.86, 0.615) (0.88, 0.611) (0.9, 0.603) (0.92, 0.598) (0.94, 0.597) (0.96, 0.595) (0.98, 0.592) (1, 0.590) 
};
\addplot[name path=GCSpS_theory_lower, draw=none] coordinates {
(0.0, 0.763) (0.1, 0.760) (0.3, 0.740) (0.5, 0.698) (0.7, 0.660) (0.72, 0.655) (0.74, 0.648) (0.76, 0.643) (0.78, 0.641) (0.8, 0.636) (0.82, 0.626) (0.84, 0.622) (0.86, 0.613) (0.88, 0.608) (0.9, 0.601) (0.92, 0.596) (0.94, 0.595) (0.96, 0.593) (0.98, 0.590) (1, 0.588) 
};
\addplot[GCSpS_theory_style, opacity=0.2] fill between[of=GCSpS_theory_upper and GCSpS_theory_lower];

\addplot[GCSpS_theory_style, mark indices={1,2,3,4,5,13,20}] coordinates {
(0.0, 0.764) (0.1, 0.761) (0.3, 0.742) (0.5, 0.699) (0.7, 0.661) (0.72, 0.656) (0.74, 0.650) (0.76, 0.644) (0.78, 0.643) (0.8, 0.637) (0.82, 0.628) (0.84, 0.623) (0.86, 0.614) (0.88, 0.610) (0.9, 0.602) (0.92, 0.597) (0.94, 0.596) (0.96, 0.594) (0.98, 0.591) (1, 0.589) 
};

\addplot[name path=wr_optimal_theory_upper, draw=none] coordinates {
(0.0, 0.660) (0.1, 0.667) (0.3, 0.669) (0.5, 0.638) (0.7, 0.609) (0.72, 0.606) (0.74, 0.601) (0.76, 0.599) (0.78, 0.599) (0.8, 0.596) (0.82, 0.589) (0.84, 0.588) (0.86, 0.584) (0.88, 0.583) (0.9, 0.581) (0.92, 0.585) (0.94, 0.590) (0.96, 0.596) (0.98, 0.601) (1, 0.610) 
};
\addplot[name path=wr_optimal_theory_lower, draw=none] coordinates {
(0.0, 0.658) (0.1, 0.665) (0.3, 0.666) (0.5, 0.636) (0.7, 0.607) (0.72, 0.604) (0.74, 0.599) (0.76, 0.596) (0.78, 0.597) (0.8, 0.594) (0.82, 0.587) (0.84, 0.586) (0.86, 0.582) (0.88, 0.581) (0.9, 0.579) (0.92, 0.583) (0.94, 0.588) (0.96, 0.595) (0.98, 0.599) (1, 0.608) 
};
\addplot[wr_optimal_theory_style, opacity=0.2] fill between[of=wr_optimal_theory_upper and wr_optimal_theory_lower];

\addplot[wr_optimal_theory_style, mark indices={1,2,3,4,5,13,20}] coordinates {
(0.0, 0.659) (0.1, 0.666) (0.3, 0.668) (0.5, 0.637) (0.7, 0.608) (0.72, 0.605) (0.74, 0.600) (0.76, 0.597) (0.78, 0.598) (0.8, 0.595) (0.82, 0.588) (0.84, 0.587) (0.86, 0.583) (0.88, 0.582) (0.9, 0.580) (0.92, 0.584) (0.94, 0.589) (0.96, 0.595) (0.98, 0.600) (1, 0.609) 
};

\addplot[name path=wor_optimal_theory_upper, draw=none] coordinates {
(0.0, 0.765) (0.1, 0.762) (0.3, 0.739) (0.5, 0.686) (0.7, 0.635) (0.72, 0.630) (0.74, 0.623) (0.76, 0.619) (0.78, 0.618) (0.8, 0.613) (0.82, 0.606) (0.84, 0.603) (0.86, 0.597) (0.88, 0.595) (0.9, 0.593) (0.92, 0.596) (0.94, 0.601) (0.96, 0.606) (0.98, 0.610) (1, 0.619) 
};
\addplot[name path=wor_optimal_theory_lower, draw=none] coordinates {
(0.0, 0.763) (0.1, 0.760) (0.3, 0.736) (0.5, 0.684) (0.7, 0.633) (0.72, 0.627) (0.74, 0.621) (0.76, 0.617) (0.78, 0.615) (0.8, 0.611) (0.82, 0.603) (0.84, 0.601) (0.86, 0.595) (0.88, 0.593) (0.9, 0.591) (0.92, 0.594) (0.94, 0.599) (0.96, 0.604) (0.98, 0.609) (1, 0.618) 
};
\addplot[wor_optimal_theory_style, opacity=0.2] fill between[of=wor_optimal_theory_upper and wor_optimal_theory_lower];

\addplot[wor_optimal_theory_style, mark indices={1,2,3,4,5,13,20}] coordinates {
(0.0, 0.764) (0.1, 0.761) (0.3, 0.738) (0.5, 0.685) (0.7, 0.634) (0.72, 0.628) (0.74, 0.622) (0.76, 0.618) (0.78, 0.616) (0.8, 0.612) (0.82, 0.604) (0.84, 0.602) (0.86, 0.596) (0.88, 0.594) (0.9, 0.592) (0.92, 0.595) (0.94, 0.600) (0.96, 0.605) (0.98, 0.609) (1, 0.618) 
};

 \end{axis}
\end{tikzpicture}
\caption{WMT'14 De-En}
\end{subfigure}
\hspace{-2.2em} \begin{subfigure}[b]{0.36\textwidth} \begin{tikzpicture}
\begin{axis}[
    width=\textwidth,
    height=0.8\textwidth,
    xlabel={Temperatures},
    xlabel style={font=\scriptsize, yshift=0.5em}, ylabel style={font=\scriptsize, yshift=-1.5em}, ylabel={$\alpha$},
    xtick={0.1, 0.3, 0.5, 0.7, 0.9},
    xmin=-0.1, xmax=1.1,
xticklabel style={font=\scriptsize},
    yticklabel style={font=\scriptsize},
    grid=major,
    legend style={font=\scriptsize, at={(1.05,1)}, anchor=north west},
]
\addplot[name path=K_Seq_verify_upper, draw=none] coordinates {
(0.0, 0.630) (0.1, 0.646) (0.3, 0.655) (0.5, 0.655) (0.7, 0.661) (0.72, 0.664) (0.74, 0.663) (0.76, 0.665) (0.78, 0.667) (0.8, 0.666) (0.82, 0.667) (0.84, 0.666) (0.86, 0.667) (0.88, 0.669) (0.9, 0.665) (0.92, 0.666) (0.94, 0.668) (0.96, 0.671) (0.98, 0.672) (1, 0.673) 
};
\addplot[name path=K_Seq_verify_lower, draw=none] coordinates {
(0.0, 0.628) (0.1, 0.643) (0.3, 0.652) (0.5, 0.653) (0.7, 0.659) (0.72, 0.661) (0.74, 0.661) (0.76, 0.662) (0.78, 0.664) (0.8, 0.663) (0.82, 0.664) (0.84, 0.663) (0.86, 0.664) (0.88, 0.666) (0.9, 0.663) (0.92, 0.664) (0.94, 0.665) (0.96, 0.669) (0.98, 0.670) (1, 0.670) 
};
\addplot[K_Seq_verify_style, opacity=0.2] fill between[of=K_Seq_verify_upper and K_Seq_verify_lower];

\addplot[K_Seq_verify_style, mark indices={1,2,3,4,5,13,20}] coordinates {
(0.0, 0.629) (0.1, 0.645) (0.3, 0.653) (0.5, 0.654) (0.7, 0.660) (0.72, 0.663) (0.74, 0.662) (0.76, 0.663) (0.78, 0.665) (0.8, 0.665) (0.82, 0.665) (0.84, 0.664) (0.86, 0.665) (0.88, 0.667) (0.9, 0.664) (0.92, 0.665) (0.94, 0.667) (0.96, 0.670) (0.98, 0.671) (1, 0.672) 
};

\addplot[name path=GCSpS_verify_upper, draw=none] coordinates {
(0.0, 0.797) (0.1, 0.793) (0.3, 0.772) (0.5, 0.749) (0.7, 0.734) (0.72, 0.734) (0.74, 0.728) (0.76, 0.728) (0.78, 0.727) (0.8, 0.723) (0.82, 0.719) (0.84, 0.715) (0.86, 0.712) (0.88, 0.710) (0.9, 0.704) (0.92, 0.703) (0.94, 0.700) (0.96, 0.698) (0.98, 0.694) (1, 0.689) 
};
\addplot[name path=GCSpS_verify_lower, draw=none] coordinates {
(0.0, 0.794) (0.1, 0.790) (0.3, 0.769) (0.5, 0.747) (0.7, 0.731) (0.72, 0.731) (0.74, 0.726) (0.76, 0.725) (0.78, 0.724) (0.8, 0.721) (0.82, 0.716) (0.84, 0.712) (0.86, 0.710) (0.88, 0.707) (0.9, 0.702) (0.92, 0.700) (0.94, 0.697) (0.96, 0.696) (0.98, 0.691) (1, 0.686) 
};
\addplot[GCSpS_verify_style, opacity=0.2] fill between[of=GCSpS_verify_upper and GCSpS_verify_lower];

\addplot[GCSpS_verify_style, mark indices={1,2,3,4,5,13,20}] coordinates {
(0.0, 0.795) (0.1, 0.792) (0.3, 0.770) (0.5, 0.748) (0.7, 0.732) (0.72, 0.733) (0.74, 0.727) (0.76, 0.726) (0.78, 0.725) (0.8, 0.722) (0.82, 0.717) (0.84, 0.713) (0.86, 0.711) (0.88, 0.708) (0.9, 0.703) (0.92, 0.702) (0.94, 0.698) (0.96, 0.697) (0.98, 0.693) (1, 0.688) 
};

\addplot[name path=wr_recursive_verify_upper, draw=none] coordinates {
(0.0, 0.630) (0.1, 0.646) (0.3, 0.654) (0.5, 0.652) (0.7, 0.654) (0.72, 0.658) (0.74, 0.657) (0.76, 0.657) (0.78, 0.659) (0.8, 0.658) (0.82, 0.658) (0.84, 0.656) (0.86, 0.657) (0.88, 0.658) (0.9, 0.655) (0.92, 0.656) (0.94, 0.656) (0.96, 0.658) (0.98, 0.659) (1, 0.661) 
};
\addplot[name path=wr_recursive_verify_lower, draw=none] coordinates {
(0.0, 0.628) (0.1, 0.643) (0.3, 0.652) (0.5, 0.649) (0.7, 0.652) (0.72, 0.655) (0.74, 0.654) (0.76, 0.655) (0.78, 0.656) (0.8, 0.655) (0.82, 0.656) (0.84, 0.654) (0.86, 0.655) (0.88, 0.656) (0.9, 0.653) (0.92, 0.653) (0.94, 0.654) (0.96, 0.656) (0.98, 0.656) (1, 0.658) 
};
\addplot[wr_recursive_verify_style, opacity=0.2] fill between[of=wr_recursive_verify_upper and wr_recursive_verify_lower];

\addplot[wr_recursive_verify_style, mark indices={1,2,3,4,5,13,20}] coordinates {
(0.0, 0.629) (0.1, 0.645) (0.3, 0.653) (0.5, 0.650) (0.7, 0.653) (0.72, 0.656) (0.74, 0.655) (0.76, 0.656) (0.78, 0.658) (0.8, 0.656) (0.82, 0.657) (0.84, 0.655) (0.86, 0.656) (0.88, 0.657) (0.9, 0.654) (0.92, 0.654) (0.94, 0.655) (0.96, 0.657) (0.98, 0.658) (1, 0.660) 
};

\addplot[name path=wor_recursive_verify_upper, draw=none] coordinates {
(0.0, 0.797) (0.1, 0.792) (0.3, 0.767) (0.5, 0.732) (0.7, 0.706) (0.72, 0.705) (0.74, 0.701) (0.76, 0.699) (0.78, 0.698) (0.8, 0.694) (0.82, 0.691) (0.84, 0.690) (0.86, 0.689) (0.88, 0.688) (0.9, 0.682) (0.92, 0.682) (0.94, 0.680) (0.96, 0.681) (0.98, 0.683) (1, 0.681) 
};
\addplot[name path=wor_recursive_verify_lower, draw=none] coordinates {
(0.0, 0.794) (0.1, 0.790) (0.3, 0.764) (0.5, 0.730) (0.7, 0.704) (0.72, 0.703) (0.74, 0.698) (0.76, 0.697) (0.78, 0.696) (0.8, 0.692) (0.82, 0.688) (0.84, 0.688) (0.86, 0.686) (0.88, 0.685) (0.9, 0.679) (0.92, 0.679) (0.94, 0.678) (0.96, 0.679) (0.98, 0.680) (1, 0.678) 
};
\addplot[wor_recursive_verify_style, opacity=0.2] fill between[of=wor_recursive_verify_upper and wor_recursive_verify_lower];

\addplot[wor_recursive_verify_style, mark indices={1,2,3,4,5,13,20}] coordinates {
(0.0, 0.795) (0.1, 0.791) (0.3, 0.766) (0.5, 0.731) (0.7, 0.705) (0.72, 0.704) (0.74, 0.700) (0.76, 0.698) (0.78, 0.697) (0.8, 0.693) (0.82, 0.690) (0.84, 0.689) (0.86, 0.687) (0.88, 0.687) (0.9, 0.680) (0.92, 0.680) (0.94, 0.679) (0.96, 0.680) (0.98, 0.681) (1, 0.679) 
};

\addplot[name path=K_Seq_theory_upper, draw=none] coordinates {
(0.0, 0.630) (0.1, 0.645) (0.3, 0.654) (0.5, 0.654) (0.7, 0.662) (0.72, 0.665) (0.74, 0.664) (0.76, 0.664) (0.78, 0.666) (0.8, 0.665) (0.82, 0.665) (0.84, 0.665) (0.86, 0.666) (0.88, 0.667) (0.9, 0.665) (0.92, 0.667) (0.94, 0.668) (0.96, 0.670) (0.98, 0.672) (1, 0.673) 
};
\addplot[name path=K_Seq_theory_lower, draw=none] coordinates {
(0.0, 0.628) (0.1, 0.643) (0.3, 0.652) (0.5, 0.652) (0.7, 0.661) (0.72, 0.664) (0.74, 0.662) (0.76, 0.663) (0.78, 0.664) (0.8, 0.663) (0.82, 0.664) (0.84, 0.663) (0.86, 0.664) (0.88, 0.666) (0.9, 0.664) (0.92, 0.665) (0.94, 0.666) (0.96, 0.668) (0.98, 0.670) (1, 0.672) 
};
\addplot[K_Seq_theory_style, opacity=0.2] fill between[of=K_Seq_theory_upper and K_Seq_theory_lower];

\addplot[K_Seq_theory_style, mark indices={1,2,3,4,5,13,20}] coordinates {
(0.0, 0.629) (0.1, 0.644) (0.3, 0.653) (0.5, 0.653) (0.7, 0.662) (0.72, 0.665) (0.74, 0.663) (0.76, 0.664) (0.78, 0.665) (0.8, 0.664) (0.82, 0.665) (0.84, 0.664) (0.86, 0.665) (0.88, 0.666) (0.9, 0.664) (0.92, 0.666) (0.94, 0.667) (0.96, 0.669) (0.98, 0.671) (1, 0.672) 
};

\addplot[name path=wr_optimal_theory_upper, draw=none] coordinates {
(0.0, 0.630) (0.1, 0.645) (0.3, 0.655) (0.5, 0.660) (0.7, 0.676) (0.72, 0.679) (0.74, 0.678) (0.76, 0.680) (0.78, 0.682) (0.8, 0.682) (0.82, 0.683) (0.84, 0.684) (0.86, 0.686) (0.88, 0.688) (0.9, 0.687) (0.92, 0.690) (0.94, 0.692) (0.96, 0.695) (0.98, 0.698) (1, 0.700) 
};
\addplot[name path=wr_optimal_theory_lower, draw=none] coordinates {
(0.0, 0.628) (0.1, 0.643) (0.3, 0.653) (0.5, 0.658) (0.7, 0.674) (0.72, 0.677) (0.74, 0.677) (0.76, 0.678) (0.78, 0.680) (0.8, 0.681) (0.82, 0.682) (0.84, 0.682) (0.86, 0.684) (0.88, 0.686) (0.9, 0.685) (0.92, 0.688) (0.94, 0.690) (0.96, 0.693) (0.98, 0.696) (1, 0.699) 
};
\addplot[wr_optimal_theory_style, opacity=0.2] fill between[of=wr_optimal_theory_upper and wr_optimal_theory_lower];

\addplot[wr_optimal_theory_style, mark indices={1,2,3,4,5,13,20}] coordinates {
(0.0, 0.629) (0.1, 0.644) (0.3, 0.654) (0.5, 0.659) (0.7, 0.675) (0.72, 0.678) (0.74, 0.678) (0.76, 0.679) (0.78, 0.681) (0.8, 0.681) (0.82, 0.683) (0.84, 0.683) (0.86, 0.685) (0.88, 0.687) (0.9, 0.686) (0.92, 0.689) (0.94, 0.691) (0.96, 0.694) (0.98, 0.697) (1, 0.700) 
};

\addplot[name path=GCSpS_theory_upper, draw=none] coordinates {
(0.0, 0.797) (0.1, 0.793) (0.3, 0.771) (0.5, 0.749) (0.7, 0.732) (0.72, 0.732) (0.74, 0.728) (0.76, 0.726) (0.78, 0.725) (0.8, 0.721) (0.82, 0.718) (0.84, 0.715) (0.86, 0.712) (0.88, 0.710) (0.9, 0.704) (0.92, 0.703) (0.94, 0.699) (0.96, 0.697) (0.98, 0.694) (1, 0.691) 
};
\addplot[name path=GCSpS_theory_lower, draw=none] coordinates {
(0.0, 0.794) (0.1, 0.790) (0.3, 0.769) (0.5, 0.747) (0.7, 0.730) (0.72, 0.730) (0.74, 0.726) (0.76, 0.725) (0.78, 0.723) (0.8, 0.719) (0.82, 0.716) (0.84, 0.713) (0.86, 0.710) (0.88, 0.708) (0.9, 0.702) (0.92, 0.701) (0.94, 0.697) (0.96, 0.695) (0.98, 0.692) (1, 0.689) 
};
\addplot[GCSpS_theory_style, opacity=0.2] fill between[of=GCSpS_theory_upper and GCSpS_theory_lower];

\addplot[GCSpS_theory_style, mark indices={1,2,3,4,5,13,20}] coordinates {
(0.0, 0.795) (0.1, 0.791) (0.3, 0.770) (0.5, 0.748) (0.7, 0.731) (0.72, 0.731) (0.74, 0.727) (0.76, 0.725) (0.78, 0.724) (0.8, 0.720) (0.82, 0.717) (0.84, 0.714) (0.86, 0.711) (0.88, 0.709) (0.9, 0.703) (0.92, 0.702) (0.94, 0.698) (0.96, 0.696) (0.98, 0.693) (1, 0.690) 
};

\addplot[name path=wor_optimal_theory_upper, draw=none] coordinates {
(0.0, 0.797) (0.1, 0.792) (0.3, 0.767) (0.5, 0.737) (0.7, 0.719) (0.72, 0.720) (0.74, 0.717) (0.76, 0.716) (0.78, 0.716) (0.8, 0.714) (0.82, 0.713) (0.84, 0.711) (0.86, 0.711) (0.88, 0.712) (0.9, 0.709) (0.92, 0.710) (0.94, 0.711) (0.96, 0.713) (0.98, 0.715) (1, 0.717) 
};
\addplot[name path=wor_optimal_theory_lower, draw=none] coordinates {
(0.0, 0.794) (0.1, 0.790) (0.3, 0.765) (0.5, 0.735) (0.7, 0.717) (0.72, 0.718) (0.74, 0.715) (0.76, 0.714) (0.78, 0.714) (0.8, 0.712) (0.82, 0.711) (0.84, 0.710) (0.86, 0.710) (0.88, 0.710) (0.9, 0.707) (0.92, 0.709) (0.94, 0.710) (0.96, 0.712) (0.98, 0.714) (1, 0.715) 
};
\addplot[wor_optimal_theory_style, opacity=0.2] fill between[of=wor_optimal_theory_upper and wor_optimal_theory_lower];

\addplot[wor_optimal_theory_style, mark indices={1,2,3,4,5,13,20}] coordinates {
(0.0, 0.795) (0.1, 0.791) (0.3, 0.766) (0.5, 0.736) (0.7, 0.718) (0.72, 0.719) (0.74, 0.716) (0.76, 0.715) (0.78, 0.715) (0.8, 0.713) (0.82, 0.712) (0.84, 0.710) (0.86, 0.710) (0.88, 0.711) (0.9, 0.708) (0.92, 0.710) (0.94, 0.711) (0.96, 0.712) (0.98, 0.714) (1, 0.716) 
};

 \end{axis}
\end{tikzpicture}
\caption{CNN-DailyMail}
\end{subfigure}
\hspace{-2.2em} \begin{subfigure}[b]{0.36\textwidth} \begin{tikzpicture}
\begin{axis}[
    width=\textwidth,
    height=0.8\textwidth,
    xlabel={Temperatures},
    xlabel style={font=\scriptsize, yshift=0.5em}, ylabel style={font=\scriptsize, yshift=-1.5em}, ylabel={$\alpha$},
    xtick={0.1, 0.3, 0.5, 0.7, 0.9},
    xmin=-0.1, xmax=1.1,
xticklabel style={font=\scriptsize},
    yticklabel style={font=\scriptsize},
    grid=major,
    legend style={font=\scriptsize, at={(1.05,1)}, anchor=north west},
]
\addplot[name path=wr_recursive_verify_upper, draw=none] coordinates {
(0.0, 0.788) (0.1, 0.795) (0.3, 0.795) (0.5, 0.767) (0.7, 0.717) (0.72, 0.715) (0.74, 0.708) (0.76, 0.701) (0.78, 0.696) (0.8, 0.695) (0.82, 0.687) (0.84, 0.685) (0.86, 0.679) (0.88, 0.676) (0.9, 0.671) (0.92, 0.668) (0.94, 0.665) (0.96, 0.665) (0.98, 0.664) (1, 0.665) 
};
\addplot[name path=wr_recursive_verify_lower, draw=none] coordinates {
(0.0, 0.786) (0.1, 0.793) (0.3, 0.792) (0.5, 0.764) (0.7, 0.714) (0.72, 0.713) (0.74, 0.706) (0.76, 0.699) (0.78, 0.694) (0.8, 0.692) (0.82, 0.684) (0.84, 0.682) (0.86, 0.676) (0.88, 0.673) (0.9, 0.668) (0.92, 0.666) (0.94, 0.662) (0.96, 0.663) (0.98, 0.661) (1, 0.662) 
};
\addplot[wr_recursive_verify_style, opacity=0.2] fill between[of=wr_recursive_verify_upper and wr_recursive_verify_lower];

\addplot[wr_recursive_verify_style, mark indices={1,2,3,4,5,13,20}] coordinates {
(0.0, 0.787) (0.1, 0.794) (0.3, 0.793) (0.5, 0.766) (0.7, 0.716) (0.72, 0.714) (0.74, 0.707) (0.76, 0.700) (0.78, 0.695) (0.8, 0.693) (0.82, 0.686) (0.84, 0.683) (0.86, 0.678) (0.88, 0.674) (0.9, 0.670) (0.92, 0.667) (0.94, 0.664) (0.96, 0.664) (0.98, 0.663) (1, 0.664) 
};

\addplot[name path=wor_recursive_verify_upper, draw=none] coordinates {
(0.0, 0.884) (0.1, 0.882) (0.3, 0.865) (0.5, 0.824) (0.7, 0.758) (0.72, 0.754) (0.74, 0.745) (0.76, 0.740) (0.78, 0.732) (0.8, 0.728) (0.82, 0.717) (0.84, 0.714) (0.86, 0.709) (0.88, 0.704) (0.9, 0.697) (0.92, 0.691) (0.94, 0.686) (0.96, 0.685) (0.98, 0.681) (1, 0.682) 
};
\addplot[name path=wor_recursive_verify_lower, draw=none] coordinates {
(0.0, 0.882) (0.1, 0.880) (0.3, 0.863) (0.5, 0.822) (0.7, 0.756) (0.72, 0.752) (0.74, 0.742) (0.76, 0.737) (0.78, 0.729) (0.8, 0.725) (0.82, 0.715) (0.84, 0.712) (0.86, 0.707) (0.88, 0.701) (0.9, 0.695) (0.92, 0.688) (0.94, 0.683) (0.96, 0.682) (0.98, 0.679) (1, 0.679) 
};
\addplot[wor_recursive_verify_style, opacity=0.2] fill between[of=wor_recursive_verify_upper and wor_recursive_verify_lower];

\addplot[wor_recursive_verify_style, mark indices={1,2,3,4,5,13,20}] coordinates {
(0.0, 0.883) (0.1, 0.881) (0.3, 0.864) (0.5, 0.823) (0.7, 0.757) (0.72, 0.753) (0.74, 0.743) (0.76, 0.739) (0.78, 0.731) (0.8, 0.726) (0.82, 0.716) (0.84, 0.713) (0.86, 0.708) (0.88, 0.702) (0.9, 0.696) (0.92, 0.689) (0.94, 0.684) (0.96, 0.683) (0.98, 0.680) (1, 0.680) 
};

\addplot[name path=GCSpS_verify_upper, draw=none] coordinates {
(0.0, 0.884) (0.1, 0.882) (0.3, 0.868) (0.5, 0.836) (0.7, 0.785) (0.72, 0.782) (0.74, 0.774) (0.76, 0.770) (0.78, 0.762) (0.8, 0.758) (0.82, 0.748) (0.84, 0.743) (0.86, 0.739) (0.88, 0.730) (0.9, 0.723) (0.92, 0.713) (0.94, 0.704) (0.96, 0.700) (0.98, 0.694) (1, 0.688) 
};
\addplot[name path=GCSpS_verify_lower, draw=none] coordinates {
(0.0, 0.882) (0.1, 0.880) (0.3, 0.866) (0.5, 0.834) (0.7, 0.782) (0.72, 0.780) (0.74, 0.772) (0.76, 0.767) (0.78, 0.760) (0.8, 0.755) (0.82, 0.745) (0.84, 0.741) (0.86, 0.736) (0.88, 0.727) (0.9, 0.720) (0.92, 0.710) (0.94, 0.702) (0.96, 0.698) (0.98, 0.692) (1, 0.686) 
};
\addplot[GCSpS_verify_style, opacity=0.2] fill between[of=GCSpS_verify_upper and GCSpS_verify_lower];

\addplot[GCSpS_verify_style, mark indices={1,2,3,4,5,13,20}] coordinates {
(0.0, 0.883) (0.1, 0.881) (0.3, 0.867) (0.5, 0.835) (0.7, 0.784) (0.72, 0.781) (0.74, 0.773) (0.76, 0.769) (0.78, 0.761) (0.8, 0.757) (0.82, 0.746) (0.84, 0.742) (0.86, 0.738) (0.88, 0.729) (0.9, 0.722) (0.92, 0.712) (0.94, 0.703) (0.96, 0.699) (0.98, 0.693) (1, 0.687) 
};

\addplot[name path=K_Seq_verify_upper, draw=none] coordinates {
(0.0, 0.788) (0.1, 0.795) (0.3, 0.795) (0.5, 0.769) (0.7, 0.720) (0.72, 0.720) (0.74, 0.713) (0.76, 0.708) (0.78, 0.703) (0.8, 0.701) (0.82, 0.695) (0.84, 0.694) (0.86, 0.690) (0.88, 0.687) (0.9, 0.683) (0.92, 0.681) (0.94, 0.679) (0.96, 0.679) (0.98, 0.678) (1, 0.681) 
};
\addplot[name path=K_Seq_verify_lower, draw=none] coordinates {
(0.0, 0.786) (0.1, 0.793) (0.3, 0.792) (0.5, 0.766) (0.7, 0.718) (0.72, 0.717) (0.74, 0.711) (0.76, 0.705) (0.78, 0.700) (0.8, 0.699) (0.82, 0.692) (0.84, 0.691) (0.86, 0.687) (0.88, 0.685) (0.9, 0.681) (0.92, 0.679) (0.94, 0.676) (0.96, 0.676) (0.98, 0.675) (1, 0.678) 
};
\addplot[K_Seq_verify_style, opacity=0.2] fill between[of=K_Seq_verify_upper and K_Seq_verify_lower];

\addplot[K_Seq_verify_style, mark indices={1,2,3,4,5,13,20}] coordinates {
(0.0, 0.787) (0.1, 0.794) (0.3, 0.794) (0.5, 0.767) (0.7, 0.719) (0.72, 0.719) (0.74, 0.712) (0.76, 0.706) (0.78, 0.701) (0.8, 0.700) (0.82, 0.694) (0.84, 0.693) (0.86, 0.688) (0.88, 0.686) (0.9, 0.682) (0.92, 0.680) (0.94, 0.678) (0.96, 0.677) (0.98, 0.676) (1, 0.680) 
};

\addplot[name path=wor_optimal_theory_upper, draw=none] coordinates {
(0.0, 0.884) (0.1, 0.881) (0.3, 0.865) (0.5, 0.825) (0.7, 0.766) (0.72, 0.762) (0.74, 0.755) (0.76, 0.750) (0.78, 0.745) (0.8, 0.742) (0.82, 0.735) (0.84, 0.734) (0.86, 0.730) (0.88, 0.726) (0.9, 0.724) (0.92, 0.722) (0.94, 0.720) (0.96, 0.721) (0.98, 0.722) (1, 0.726) 
};
\addplot[name path=wor_optimal_theory_lower, draw=none] coordinates {
(0.0, 0.882) (0.1, 0.880) (0.3, 0.863) (0.5, 0.823) (0.7, 0.764) (0.72, 0.761) (0.74, 0.753) (0.76, 0.749) (0.78, 0.743) (0.8, 0.740) (0.82, 0.733) (0.84, 0.732) (0.86, 0.729) (0.88, 0.725) (0.9, 0.722) (0.92, 0.720) (0.94, 0.718) (0.96, 0.720) (0.98, 0.721) (1, 0.724) 
};
\addplot[wor_optimal_theory_style, opacity=0.2] fill between[of=wor_optimal_theory_upper and wor_optimal_theory_lower];

\addplot[wor_optimal_theory_style, mark indices={1,2,3,4,5,13,20}] coordinates {
(0.0, 0.883) (0.1, 0.880) (0.3, 0.864) (0.5, 0.824) (0.7, 0.765) (0.72, 0.761) (0.74, 0.754) (0.76, 0.750) (0.78, 0.744) (0.8, 0.741) (0.82, 0.734) (0.84, 0.733) (0.86, 0.730) (0.88, 0.726) (0.9, 0.723) (0.92, 0.721) (0.94, 0.719) (0.96, 0.721) (0.98, 0.722) (1, 0.725) 
};

\addplot[name path=wr_optimal_theory_upper, draw=none] coordinates {
(0.0, 0.788) (0.1, 0.795) (0.3, 0.795) (0.5, 0.771) (0.7, 0.731) (0.72, 0.730) (0.74, 0.725) (0.76, 0.722) (0.78, 0.718) (0.8, 0.718) (0.82, 0.712) (0.84, 0.713) (0.86, 0.711) (0.88, 0.708) (0.9, 0.706) (0.92, 0.705) (0.94, 0.704) (0.96, 0.706) (0.98, 0.708) (1, 0.712) 
};
\addplot[name path=wr_optimal_theory_lower, draw=none] coordinates {
(0.0, 0.786) (0.1, 0.793) (0.3, 0.793) (0.5, 0.769) (0.7, 0.729) (0.72, 0.728) (0.74, 0.723) (0.76, 0.720) (0.78, 0.716) (0.8, 0.716) (0.82, 0.710) (0.84, 0.711) (0.86, 0.709) (0.88, 0.706) (0.9, 0.705) (0.92, 0.704) (0.94, 0.703) (0.96, 0.705) (0.98, 0.707) (1, 0.711) 
};
\addplot[wr_optimal_theory_style, opacity=0.2] fill between[of=wr_optimal_theory_upper and wr_optimal_theory_lower];

\addplot[wr_optimal_theory_style, mark indices={1,2,3,4,5,13,20}] coordinates {
(0.0, 0.787) (0.1, 0.794) (0.3, 0.794) (0.5, 0.770) (0.7, 0.730) (0.72, 0.729) (0.74, 0.724) (0.76, 0.721) (0.78, 0.717) (0.8, 0.717) (0.82, 0.711) (0.84, 0.712) (0.86, 0.710) (0.88, 0.707) (0.9, 0.705) (0.92, 0.704) (0.94, 0.703) (0.96, 0.706) (0.98, 0.708) (1, 0.712) 
};

\addplot[name path=GCSpS_theory_upper, draw=none] coordinates {
(0.0, 0.884) (0.1, 0.882) (0.3, 0.868) (0.5, 0.837) (0.7, 0.785) (0.72, 0.782) (0.74, 0.774) (0.76, 0.769) (0.78, 0.762) (0.8, 0.758) (0.82, 0.748) (0.84, 0.744) (0.86, 0.737) (0.88, 0.730) (0.9, 0.722) (0.92, 0.713) (0.94, 0.705) (0.96, 0.701) (0.98, 0.693) (1, 0.688) 
};
\addplot[name path=GCSpS_theory_lower, draw=none] coordinates {
(0.0, 0.882) (0.1, 0.880) (0.3, 0.866) (0.5, 0.835) (0.7, 0.783) (0.72, 0.780) (0.74, 0.772) (0.76, 0.767) (0.78, 0.760) (0.8, 0.756) (0.82, 0.746) (0.84, 0.742) (0.86, 0.735) (0.88, 0.728) (0.9, 0.720) (0.92, 0.711) (0.94, 0.704) (0.96, 0.699) (0.98, 0.692) (1, 0.687) 
};
\addplot[GCSpS_theory_style, opacity=0.2] fill between[of=GCSpS_theory_upper and GCSpS_theory_lower];

\addplot[GCSpS_theory_style, mark indices={1,2,3,4,5,13,20}] coordinates {
(0.0, 0.883) (0.1, 0.881) (0.3, 0.867) (0.5, 0.836) (0.7, 0.784) (0.72, 0.781) (0.74, 0.773) (0.76, 0.768) (0.78, 0.761) (0.8, 0.757) (0.82, 0.747) (0.84, 0.743) (0.86, 0.736) (0.88, 0.729) (0.9, 0.721) (0.92, 0.712) (0.94, 0.704) (0.96, 0.700) (0.98, 0.693) (1, 0.688) 
};

\addplot[name path=K_Seq_theory_upper, draw=none] coordinates {
(0.0, 0.788) (0.1, 0.795) (0.3, 0.794) (0.5, 0.768) (0.7, 0.721) (0.72, 0.719) (0.74, 0.713) (0.76, 0.709) (0.78, 0.703) (0.8, 0.702) (0.82, 0.695) (0.84, 0.694) (0.86, 0.691) (0.88, 0.687) (0.9, 0.683) (0.92, 0.680) (0.94, 0.678) (0.96, 0.678) (0.98, 0.679) (1, 0.681) 
};
\addplot[name path=K_Seq_theory_lower, draw=none] coordinates {
(0.0, 0.786) (0.1, 0.793) (0.3, 0.792) (0.5, 0.766) (0.7, 0.719) (0.72, 0.717) (0.74, 0.711) (0.76, 0.707) (0.78, 0.702) (0.8, 0.700) (0.82, 0.693) (0.84, 0.692) (0.86, 0.689) (0.88, 0.685) (0.9, 0.681) (0.92, 0.679) (0.94, 0.676) (0.96, 0.677) (0.98, 0.677) (1, 0.679) 
};
\addplot[K_Seq_theory_style, opacity=0.2] fill between[of=K_Seq_theory_upper and K_Seq_theory_lower];

\addplot[K_Seq_theory_style, mark indices={1,2,3,4,5,13,20}] coordinates {
(0.0, 0.787) (0.1, 0.794) (0.3, 0.793) (0.5, 0.767) (0.7, 0.720) (0.72, 0.718) (0.74, 0.712) (0.76, 0.708) (0.78, 0.702) (0.8, 0.701) (0.82, 0.694) (0.84, 0.693) (0.86, 0.690) (0.88, 0.686) (0.9, 0.682) (0.92, 0.680) (0.94, 0.677) (0.96, 0.678) (0.98, 0.678) (1, 0.680) 
};

 \end{axis}
\end{tikzpicture}
\caption{Alpaca}
\end{subfigure}
\crvspace{-10pt}
\caption{Comparison of acceptance rate $\alpha$ for different temperatures across datasets.}\label{fig:ab_temperature}
\end{figure}
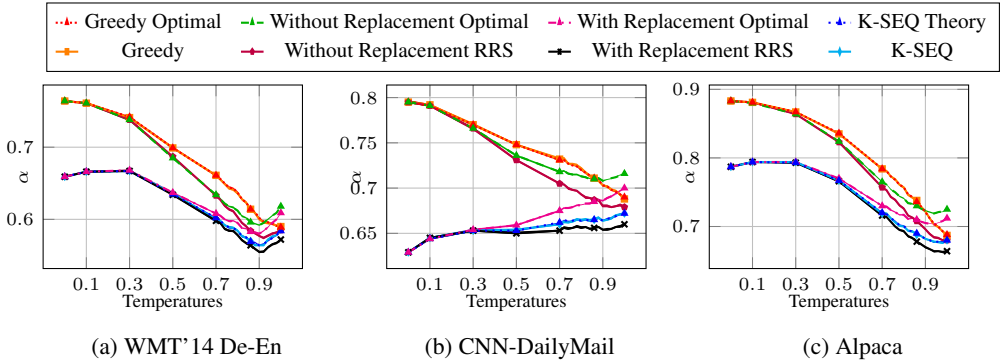 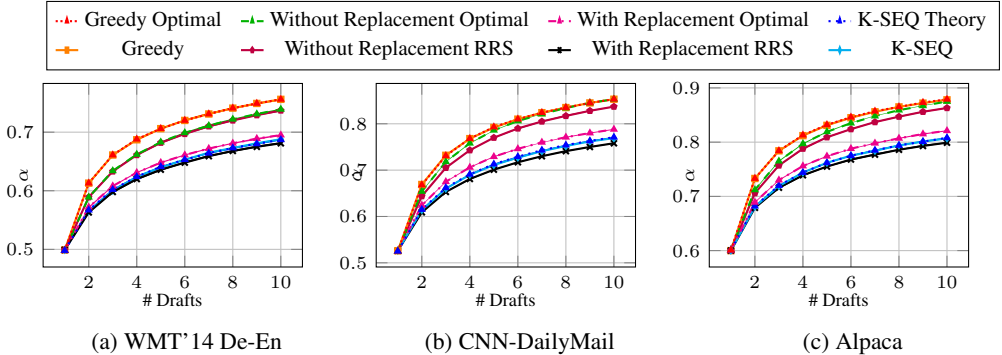
\begin{figure}[ht]
\pgfplotsset{
    GCSpS_theory_style/.style={
        color=red, thick, densely dotted, mark=triangle*, mark options={solid, scale=0.5}
    },
    wor_optimal_theory_style/.style={
        color=green!70!black, thick, dashed, mark=triangle*, mark options={solid, scale=0.5}
    },
    wr_optimal_theory_style/.style={
        color=magenta, thick, dash dot, mark=triangle*, mark options={solid, scale=0.5}
    },
    K_Seq_theory_style/.style={
        color=blue, thick, dotted, mark=triangle*, mark options={solid, scale=0.5}
    },
    GCSpS_verify_style/.style={
        color=orange, thick, solid, mark=square*, mark options={solid, scale=0.5}
    },
    wor_recursive_verify_style/.style={
        color=purple, thick, thick, mark=pentagon*, mark options={solid, scale=0.5}
    },
    wr_recursive_verify_style/.style={
        color=black, thick, solid, mark=x, mark options={solid, scale=0.8}
    },
    K_Seq_verify_style/.style={
        color=cyan, thick, solid, mark=diamond*, mark options={solid, scale=0.5}
    }
}
\centering
\begin{tikzpicture}
    \begin{axis}[
        hide axis, xmin=0, xmax=1, ymin=0, ymax=1, legend columns=4, legend style={/tikz/every even column/.append style={column sep=0.5em}}, width=10cm, height=2cm, legend image post style={xscale=0.5}, ]
   
    \addplot[GCSpS_theory_style] coordinates {(0,0)};
    \addlegendentry{\scalebox{0.8}{Greedy Optimal}}
    
    \addplot[wor_optimal_theory_style] coordinates {(0,0)};
    \addlegendentry{\scalebox{0.8}{CPS Optimal}}
    
    \addplot[wr_optimal_theory_style] coordinates {(0,0)};
    \addlegendentry{\scalebox{0.8}{With Replacement Optimal}}

    \addplot[K_Seq_theory_style] coordinates {(0,0)};
    \addlegendentry{\scalebox{0.8}{K-SEQ Theory}}

    \addplot[GCSpS_verify_style] coordinates {(0,0)};
    \addlegendentry{\scalebox{0.8}{Greedy}}
    
    \addplot[wor_recursive_verify_style] coordinates {(0,0)};
    \addlegendentry{\scalebox{0.8}{Successive RRS}}
     
    \addplot[wr_recursive_verify_style] coordinates {(0,0)};
    \addlegendentry{\scalebox{0.8}{With Replacement RRS}}
    
    \addplot[K_Seq_verify_style] coordinates {(0,0)};
    \addlegendentry{\scalebox{0.8}{K-SEQ}}
    
    \end{axis}
\end{tikzpicture}
 
\begin{subfigure}[b]{0.36\textwidth} \begin{tikzpicture}
\begin{axis}[
    width=\textwidth,
    height=0.8\textwidth,
    xlabel={\# Drafts},
    xlabel style={font=\scriptsize, yshift=0.5em}, ylabel style={font=\scriptsize, yshift=-1.5em}, ylabel={$\alpha$},
    xtick={2, 4, 6, 8, 10},
xticklabel style={font=\scriptsize},
    yticklabel style={font=\scriptsize},
    grid=major,
    legend style={font=\scriptsize, at={(1.05,1)}, anchor=north west},
]
\addplot[name path=GCSpS_verify_upper, draw=none] coordinates {
(1, 0.501) (2, 0.614) (3, 0.662) (4, 0.689) (5, 0.707) (6, 0.722) (7, 0.733) (8, 0.742) (9, 0.750) (10, 0.757) 
};
\addplot[name path=GCSpS_verify_lower, draw=none] coordinates {
(1, 0.498) (2, 0.611) (3, 0.659) (4, 0.686) (5, 0.705) (6, 0.719) (7, 0.730) (8, 0.739) (9, 0.747) (10, 0.754) 
};
\addplot[GCSpS_verify_style, opacity=0.2] fill between[of=GCSpS_verify_upper and GCSpS_verify_lower];

\addplot[GCSpS_verify_style] coordinates {
(1, 0.499) (2, 0.613) (3, 0.661) (4, 0.687) (5, 0.706) (6, 0.720) (7, 0.731) (8, 0.741) (9, 0.749) (10, 0.756) 
};

\addplot[name path=K_Seq_verify_upper, draw=none] coordinates {
(1, 0.501) (2, 0.566) (3, 0.601) (4, 0.625) (5, 0.641) (6, 0.654) (7, 0.666) (8, 0.674) (9, 0.682) (10, 0.689) 
};
\addplot[name path=K_Seq_verify_lower, draw=none] coordinates {
(1, 0.498) (2, 0.564) (3, 0.599) (4, 0.622) (5, 0.638) (6, 0.652) (7, 0.663) (8, 0.672) (9, 0.679) (10, 0.686) 
};
\addplot[K_Seq_verify_style, opacity=0.2] fill between[of=K_Seq_verify_upper and K_Seq_verify_lower];

\addplot[K_Seq_verify_style] coordinates {
(1, 0.499) (2, 0.565) (3, 0.600) (4, 0.624) (5, 0.640) (6, 0.653) (7, 0.665) (8, 0.673) (9, 0.681) (10, 0.688) 
};

\addplot[name path=wr_recursive_verify_upper, draw=none] coordinates {
(1, 0.501) (2, 0.565) (3, 0.599) (4, 0.621) (5, 0.637) (6, 0.650) (7, 0.661) (8, 0.669) (9, 0.677) (10, 0.683) 
};
\addplot[name path=wr_recursive_verify_lower, draw=none] coordinates {
(1, 0.498) (2, 0.562) (3, 0.596) (4, 0.618) (5, 0.634) (6, 0.647) (7, 0.658) (8, 0.666) (9, 0.674) (10, 0.680) 
};
\addplot[wr_recursive_verify_style, opacity=0.2] fill between[of=wr_recursive_verify_upper and wr_recursive_verify_lower];

\addplot[wr_recursive_verify_style] coordinates {
(1, 0.499) (2, 0.563) (3, 0.598) (4, 0.620) (5, 0.636) (6, 0.648) (7, 0.659) (8, 0.668) (9, 0.675) (10, 0.681) 
};

\addplot[name path=wor_recursive_verify_upper, draw=none] coordinates {
(1, 0.501) (2, 0.591) (3, 0.634) (4, 0.663) (5, 0.683) (6, 0.699) (7, 0.711) (8, 0.722) (9, 0.730) (10, 0.738) 
};
\addplot[name path=wor_recursive_verify_lower, draw=none] coordinates {
(1, 0.498) (2, 0.588) (3, 0.632) (4, 0.660) (5, 0.680) (6, 0.696) (7, 0.708) (8, 0.719) (9, 0.728) (10, 0.735) 
};
\addplot[wor_recursive_verify_style, opacity=0.2] fill between[of=wor_recursive_verify_upper and wor_recursive_verify_lower];

\addplot[wor_recursive_verify_style] coordinates {
(1, 0.499) (2, 0.589) (3, 0.633) (4, 0.661) (5, 0.682) (6, 0.697) (7, 0.710) (8, 0.720) (9, 0.729) (10, 0.737) 
};

\addplot[name path=wor_optimal_theory_upper, draw=none] coordinates {
(1, 0.499) (2, 0.590) (3, 0.635) (4, 0.663) (5, 0.684) (6, 0.700) (7, 0.713) (8, 0.723) (9, 0.732) (10, 0.740) 
};
\addplot[name path=wor_optimal_theory_lower, draw=none] coordinates {
(1, 0.496) (2, 0.588) (3, 0.633) (4, 0.661) (5, 0.682) (6, 0.698) (7, 0.710) (8, 0.721) (9, 0.730) (10, 0.738) 
};
\addplot[wor_optimal_theory_style, opacity=0.2] fill between[of=wor_optimal_theory_upper and wor_optimal_theory_lower];

\addplot[wor_optimal_theory_style] coordinates {
(1, 0.498) (2, 0.589) (3, 0.634) (4, 0.662) (5, 0.683) (6, 0.699) (7, 0.711) (8, 0.722) (9, 0.731) (10, 0.739) 
};

\addplot[name path=wr_optimal_theory_upper, draw=none] coordinates {
(1, 0.499) (2, 0.573) (3, 0.609) (4, 0.632) (5, 0.649) (6, 0.662) (7, 0.673) (8, 0.682) (9, 0.690) (10, 0.696) 
};
\addplot[name path=wr_optimal_theory_lower, draw=none] coordinates {
(1, 0.496) (2, 0.570) (3, 0.607) (4, 0.630) (5, 0.647) (6, 0.660) (7, 0.671) (8, 0.680) (9, 0.687) (10, 0.694) 
};
\addplot[wr_optimal_theory_style, opacity=0.2] fill between[of=wr_optimal_theory_upper and wr_optimal_theory_lower];

\addplot[wr_optimal_theory_style] coordinates {
(1, 0.498) (2, 0.572) (3, 0.608) (4, 0.631) (5, 0.648) (6, 0.661) (7, 0.672) (8, 0.681) (9, 0.689) (10, 0.695) 
};

\addplot[name path=GCSpS_theory_upper, draw=none] coordinates {
(1, 0.499) (2, 0.614) (3, 0.662) (4, 0.689) (5, 0.707) (6, 0.721) (7, 0.733) (8, 0.742) (9, 0.750) (10, 0.757) 
};
\addplot[name path=GCSpS_theory_lower, draw=none] coordinates {
(1, 0.496) (2, 0.611) (3, 0.660) (4, 0.686) (5, 0.705) (6, 0.719) (7, 0.730) (8, 0.740) (9, 0.748) (10, 0.755) 
};
\addplot[GCSpS_theory_style, opacity=0.2] fill between[of=GCSpS_theory_upper and GCSpS_theory_lower];

\addplot[GCSpS_theory_style] coordinates {
(1, 0.498) (2, 0.613) (3, 0.661) (4, 0.688) (5, 0.706) (6, 0.720) (7, 0.731) (8, 0.741) (9, 0.749) (10, 0.756) 
};

\addplot[name path=K_Seq_theory_upper, draw=none] coordinates {
(1, 0.499) (2, 0.568) (3, 0.603) (4, 0.625) (5, 0.642) (6, 0.654) (7, 0.665) (8, 0.674) (9, 0.681) (10, 0.688) 
};
\addplot[name path=K_Seq_theory_lower, draw=none] coordinates {
(1, 0.496) (2, 0.566) (3, 0.601) (4, 0.623) (5, 0.639) (6, 0.652) (7, 0.663) (8, 0.671) (9, 0.679) (10, 0.686) 
};
\addplot[K_Seq_theory_style, opacity=0.2] fill between[of=K_Seq_theory_upper and K_Seq_theory_lower];

\addplot[K_Seq_theory_style] coordinates {
(1, 0.498) (2, 0.567) (3, 0.602) (4, 0.624) (5, 0.641) (6, 0.653) (7, 0.664) (8, 0.673) (9, 0.680) (10, 0.687) 
};

 \end{axis}
\end{tikzpicture}
\caption{WMT'14 De-En}
\end{subfigure}
\hspace{-2.2em} \begin{subfigure}[b]{0.36\textwidth} \begin{tikzpicture}
\begin{axis}[
    width=\textwidth,
    height=0.8\textwidth,
    xlabel={\# Drafts},
    xlabel style={font=\scriptsize, yshift=0.5em}, ylabel style={font=\scriptsize, yshift=-1.5em}, ylabel={$\alpha$},
    xtick={2, 4, 6, 8, 10},
xticklabel style={font=\scriptsize},
    yticklabel style={font=\scriptsize},
    grid=major,
    legend style={font=\scriptsize, at={(1.05,1)}, anchor=north west},
]
\addplot[name path=wor_recursive_verify_upper, draw=none] coordinates {
(1, 0.527) (2, 0.645) (3, 0.706) (4, 0.744) (5, 0.771) (6, 0.791) (7, 0.806) (8, 0.819) (9, 0.829) (10, 0.838) 
};
\addplot[name path=wor_recursive_verify_lower, draw=none] coordinates {
(1, 0.525) (2, 0.642) (3, 0.704) (4, 0.742) (5, 0.769) (6, 0.788) (7, 0.804) (8, 0.816) (9, 0.827) (10, 0.836) 
};
\addplot[wor_recursive_verify_style, opacity=0.2] fill between[of=wor_recursive_verify_upper and wor_recursive_verify_lower];

\addplot[wor_recursive_verify_style] coordinates {
(1, 0.526) (2, 0.644) (3, 0.705) (4, 0.743) (5, 0.770) (6, 0.790) (7, 0.805) (8, 0.817) (9, 0.828) (10, 0.837) 
};

\addplot[name path=wr_recursive_verify_upper, draw=none] coordinates {
(1, 0.527) (2, 0.611) (3, 0.654) (4, 0.682) (5, 0.702) (6, 0.718) (7, 0.731) (8, 0.742) (9, 0.751) (10, 0.759) 
};
\addplot[name path=wr_recursive_verify_lower, draw=none] coordinates {
(1, 0.525) (2, 0.608) (3, 0.652) (4, 0.679) (5, 0.700) (6, 0.716) (7, 0.729) (8, 0.740) (9, 0.749) (10, 0.757) 
};
\addplot[wr_recursive_verify_style, opacity=0.2] fill between[of=wr_recursive_verify_upper and wr_recursive_verify_lower];

\addplot[wr_recursive_verify_style] coordinates {
(1, 0.526) (2, 0.609) (3, 0.653) (4, 0.681) (5, 0.701) (6, 0.717) (7, 0.730) (8, 0.741) (9, 0.750) (10, 0.758) 
};

\addplot[name path=K_Seq_verify_upper, draw=none] coordinates {
(1, 0.527) (2, 0.615) (3, 0.661) (4, 0.691) (5, 0.713) (6, 0.728) (7, 0.742) (8, 0.754) (9, 0.763) (10, 0.770) 
};
\addplot[name path=K_Seq_verify_lower, draw=none] coordinates {
(1, 0.525) (2, 0.612) (3, 0.659) (4, 0.689) (5, 0.710) (6, 0.725) (7, 0.740) (8, 0.751) (9, 0.761) (10, 0.768) 
};
\addplot[K_Seq_verify_style, opacity=0.2] fill between[of=K_Seq_verify_upper and K_Seq_verify_lower];

\addplot[K_Seq_verify_style] coordinates {
(1, 0.526) (2, 0.614) (3, 0.660) (4, 0.690) (5, 0.711) (6, 0.727) (7, 0.741) (8, 0.752) (9, 0.762) (10, 0.769) 
};

\addplot[name path=GCSpS_verify_upper, draw=none] coordinates {
(1, 0.527) (2, 0.670) (3, 0.734) (4, 0.769) (5, 0.793) (6, 0.811) (7, 0.825) (8, 0.836) (9, 0.846) (10, 0.854) 
};
\addplot[name path=GCSpS_verify_lower, draw=none] coordinates {
(1, 0.525) (2, 0.667) (3, 0.731) (4, 0.767) (5, 0.791) (6, 0.809) (7, 0.823) (8, 0.834) (9, 0.844) (10, 0.852) 
};
\addplot[GCSpS_verify_style, opacity=0.2] fill between[of=GCSpS_verify_upper and GCSpS_verify_lower];

\addplot[GCSpS_verify_style] coordinates {
(1, 0.526) (2, 0.669) (3, 0.732) (4, 0.768) (5, 0.792) (6, 0.810) (7, 0.824) (8, 0.835) (9, 0.845) (10, 0.853) 
};

\addplot[name path=wr_optimal_theory_upper, draw=none] coordinates {
(1, 0.526) (2, 0.625) (3, 0.676) (4, 0.707) (5, 0.730) (6, 0.747) (7, 0.760) (8, 0.772) (9, 0.781) (10, 0.789) 
};
\addplot[name path=wr_optimal_theory_lower, draw=none] coordinates {
(1, 0.524) (2, 0.623) (3, 0.674) (4, 0.705) (5, 0.728) (6, 0.745) (7, 0.759) (8, 0.770) (9, 0.779) (10, 0.787) 
};
\addplot[wr_optimal_theory_style, opacity=0.2] fill between[of=wr_optimal_theory_upper and wr_optimal_theory_lower];

\addplot[wr_optimal_theory_style] coordinates {
(1, 0.525) (2, 0.624) (3, 0.675) (4, 0.706) (5, 0.729) (6, 0.746) (7, 0.760) (8, 0.771) (9, 0.780) (10, 0.788) 
};

\addplot[name path=wor_optimal_theory_upper, draw=none] coordinates {
(1, 0.526) (2, 0.655) (3, 0.719) (4, 0.759) (5, 0.786) (6, 0.807) (7, 0.822) (8, 0.835) (9, 0.845) (10, 0.854) 
};
\addplot[name path=wor_optimal_theory_lower, draw=none] coordinates {
(1, 0.524) (2, 0.653) (3, 0.717) (4, 0.757) (5, 0.785) (6, 0.805) (7, 0.821) (8, 0.833) (9, 0.844) (10, 0.853) 
};
\addplot[wor_optimal_theory_style, opacity=0.2] fill between[of=wor_optimal_theory_upper and wor_optimal_theory_lower];

\addplot[wor_optimal_theory_style] coordinates {
(1, 0.525) (2, 0.654) (3, 0.718) (4, 0.758) (5, 0.786) (6, 0.806) (7, 0.821) (8, 0.834) (9, 0.845) (10, 0.853) 
};

\addplot[name path=GCSpS_theory_upper, draw=none] coordinates {
(1, 0.526) (2, 0.669) (3, 0.732) (4, 0.769) (5, 0.793) (6, 0.811) (7, 0.825) (8, 0.836) (9, 0.846) (10, 0.854) 
};
\addplot[name path=GCSpS_theory_lower, draw=none] coordinates {
(1, 0.524) (2, 0.667) (3, 0.730) (4, 0.767) (5, 0.792) (6, 0.809) (7, 0.823) (8, 0.835) (9, 0.844) (10, 0.852) 
};
\addplot[GCSpS_theory_style, opacity=0.2] fill between[of=GCSpS_theory_upper and GCSpS_theory_lower];

\addplot[GCSpS_theory_style] coordinates {
(1, 0.525) (2, 0.668) (3, 0.731) (4, 0.768) (5, 0.793) (6, 0.810) (7, 0.824) (8, 0.835) (9, 0.845) (10, 0.853) 
};

\addplot[name path=K_Seq_theory_upper, draw=none] coordinates {
(1, 0.526) (2, 0.616) (3, 0.662) (4, 0.692) (5, 0.714) (6, 0.730) (7, 0.743) (8, 0.755) (9, 0.764) (10, 0.772) 
};
\addplot[name path=K_Seq_theory_lower, draw=none] coordinates {
(1, 0.524) (2, 0.615) (3, 0.661) (4, 0.690) (5, 0.712) (6, 0.728) (7, 0.742) (8, 0.753) (9, 0.762) (10, 0.770) 
};
\addplot[K_Seq_theory_style, opacity=0.2] fill between[of=K_Seq_theory_upper and K_Seq_theory_lower];

\addplot[K_Seq_theory_style] coordinates {
(1, 0.525) (2, 0.615) (3, 0.662) (4, 0.691) (5, 0.713) (6, 0.729) (7, 0.743) (8, 0.754) (9, 0.763) (10, 0.771) 
};

 \end{axis}
\end{tikzpicture}
\caption{CNN-DailyMail}
\end{subfigure}
\hspace{-2.2em} \begin{subfigure}[b]{0.36\textwidth} \begin{tikzpicture}
\begin{axis}[
    width=\textwidth,
    height=0.8\textwidth,
    xlabel={\# Drafts},
    xlabel style={font=\scriptsize, yshift=0.5em}, ylabel style={font=\scriptsize, yshift=-1.5em}, ylabel={$\alpha$},
    xtick={2, 4, 6, 8, 10},
xticklabel style={font=\scriptsize},
    yticklabel style={font=\scriptsize},
    grid=major,
]
\addplot[name path=GCSpS_verify_upper, draw=none] coordinates {
(1, 0.601) (2, 0.734) (3, 0.785) (4, 0.813) (5, 0.832) (6, 0.846) (7, 0.857) (8, 0.866) (9, 0.873) (10, 0.879) 
};
\addplot[name path=GCSpS_verify_lower, draw=none] coordinates {
(1, 0.598) (2, 0.732) (3, 0.782) (4, 0.811) (5, 0.830) (6, 0.844) (7, 0.855) (8, 0.864) (9, 0.871) (10, 0.877) 
};
\addplot[GCSpS_verify_style, opacity=0.2] fill between[of=GCSpS_verify_upper and GCSpS_verify_lower];

\addplot[GCSpS_verify_style] coordinates {
(1, 0.600) (2, 0.733) (3, 0.784) (4, 0.812) (5, 0.831) (6, 0.845) (7, 0.856) (8, 0.865) (9, 0.872) (10, 0.878) 
};

\addplot[name path=wr_recursive_verify_upper, draw=none] coordinates {
(1, 0.601) (2, 0.681) (3, 0.717) (4, 0.740) (5, 0.756) (6, 0.769) (7, 0.779) (8, 0.787) (9, 0.795) (10, 0.800) 
};
\addplot[name path=wr_recursive_verify_lower, draw=none] coordinates {
(1, 0.598) (2, 0.678) (3, 0.714) (4, 0.737) (5, 0.754) (6, 0.766) (7, 0.776) (8, 0.785) (9, 0.792) (10, 0.798) 
};
\addplot[wr_recursive_verify_style, opacity=0.2] fill between[of=wr_recursive_verify_upper and wr_recursive_verify_lower];

\addplot[wr_recursive_verify_style] coordinates {
(1, 0.600) (2, 0.679) (3, 0.716) (4, 0.739) (5, 0.755) (6, 0.768) (7, 0.777) (8, 0.786) (9, 0.793) (10, 0.799) 
};

\addplot[name path=wor_recursive_verify_upper, draw=none] coordinates {
(1, 0.601) (2, 0.707) (3, 0.758) (4, 0.789) (5, 0.810) (6, 0.825) (7, 0.838) (8, 0.848) (9, 0.857) (10, 0.864) 
};
\addplot[name path=wor_recursive_verify_lower, draw=none] coordinates {
(1, 0.598) (2, 0.705) (3, 0.756) (4, 0.786) (5, 0.808) (6, 0.823) (7, 0.836) (8, 0.846) (9, 0.855) (10, 0.862) 
};
\addplot[wor_recursive_verify_style, opacity=0.2] fill between[of=wor_recursive_verify_upper and wor_recursive_verify_lower];

\addplot[wor_recursive_verify_style] coordinates {
(1, 0.600) (2, 0.706) (3, 0.757) (4, 0.788) (5, 0.809) (6, 0.824) (7, 0.837) (8, 0.847) (9, 0.856) (10, 0.863) 
};

\addplot[name path=K_Seq_verify_upper, draw=none] coordinates {
(1, 0.601) (2, 0.683) (3, 0.720) (4, 0.745) (5, 0.763) (6, 0.776) (7, 0.786) (8, 0.795) (9, 0.803) (10, 0.808) 
};
\addplot[name path=K_Seq_verify_lower, draw=none] coordinates {
(1, 0.598) (2, 0.680) (3, 0.718) (4, 0.743) (5, 0.761) (6, 0.774) (7, 0.783) (8, 0.793) (9, 0.800) (10, 0.806) 
};
\addplot[K_Seq_verify_style, opacity=0.2] fill between[of=K_Seq_verify_upper and K_Seq_verify_lower];

\addplot[K_Seq_verify_style] coordinates {
(1, 0.600) (2, 0.681) (3, 0.719) (4, 0.744) (5, 0.762) (6, 0.775) (7, 0.784) (8, 0.794) (9, 0.801) (10, 0.807) 
};

\addplot[name path=wor_optimal_theory_upper, draw=none] coordinates {
(1, 0.602) (2, 0.713) (3, 0.766) (4, 0.798) (5, 0.820) (6, 0.836) (7, 0.849) (8, 0.859) (9, 0.868) (10, 0.875) 
};
\addplot[name path=wor_optimal_theory_lower, draw=none] coordinates {
(1, 0.600) (2, 0.711) (3, 0.764) (4, 0.796) (5, 0.818) (6, 0.835) (7, 0.848) (8, 0.858) (9, 0.867) (10, 0.874) 
};
\addplot[wor_optimal_theory_style, opacity=0.2] fill between[of=wor_optimal_theory_upper and wor_optimal_theory_lower];

\addplot[wor_optimal_theory_style] coordinates {
(1, 0.601) (2, 0.712) (3, 0.765) (4, 0.797) (5, 0.819) (6, 0.835) (7, 0.848) (8, 0.859) (9, 0.867) (10, 0.875) 
};

\addplot[name path=K_Seq_theory_upper, draw=none] coordinates {
(1, 0.602) (2, 0.683) (3, 0.721) (4, 0.745) (5, 0.762) (6, 0.776) (7, 0.786) (8, 0.795) (9, 0.802) (10, 0.809) 
};
\addplot[name path=K_Seq_theory_lower, draw=none] coordinates {
(1, 0.600) (2, 0.681) (3, 0.719) (4, 0.743) (5, 0.761) (6, 0.774) (7, 0.784) (8, 0.793) (9, 0.801) (10, 0.807) 
};
\addplot[K_Seq_theory_style, opacity=0.2] fill between[of=K_Seq_theory_upper and K_Seq_theory_lower];

\addplot[K_Seq_theory_style] coordinates {
(1, 0.601) (2, 0.682) (3, 0.720) (4, 0.744) (5, 0.762) (6, 0.775) (7, 0.785) (8, 0.794) (9, 0.801) (10, 0.808) 
};

\addplot[name path=wr_optimal_theory_upper, draw=none] coordinates {
(1, 0.602) (2, 0.690) (3, 0.731) (4, 0.757) (5, 0.775) (6, 0.788) (7, 0.799) (8, 0.808) (9, 0.816) (10, 0.822) 
};
\addplot[name path=wr_optimal_theory_lower, draw=none] coordinates {
(1, 0.600) (2, 0.688) (3, 0.729) (4, 0.755) (5, 0.773) (6, 0.787) (7, 0.797) (8, 0.806) (9, 0.814) (10, 0.820) 
};
\addplot[wr_optimal_theory_style, opacity=0.2] fill between[of=wr_optimal_theory_upper and wr_optimal_theory_lower];

\addplot[wr_optimal_theory_style] coordinates {
(1, 0.601) (2, 0.689) (3, 0.730) (4, 0.756) (5, 0.774) (6, 0.788) (7, 0.798) (8, 0.807) (9, 0.815) (10, 0.821) 
};

\addplot[name path=GCSpS_theory_upper, draw=none] coordinates {
(1, 0.602) (2, 0.734) (3, 0.785) (4, 0.814) (5, 0.833) (6, 0.847) (7, 0.857) (8, 0.866) (9, 0.873) (10, 0.880) 
};
\addplot[name path=GCSpS_theory_lower, draw=none] coordinates {
(1, 0.600) (2, 0.732) (3, 0.783) (4, 0.812) (5, 0.831) (6, 0.845) (7, 0.856) (8, 0.865) (9, 0.872) (10, 0.878) 
};
\addplot[GCSpS_theory_style, opacity=0.2] fill between[of=GCSpS_theory_upper and GCSpS_theory_lower];

\addplot[GCSpS_theory_style] coordinates {
(1, 0.601) (2, 0.733) (3, 0.784) (4, 0.813) (5, 0.832) (6, 0.846) (7, 0.857) (8, 0.865) (9, 0.873) (10, 0.879) 
};

 \end{axis}
\end{tikzpicture}
\caption{Alpaca}
\end{subfigure}
\crvspace{-10pt}
\caption{Comparison of acceptance rate $\alpha$ for different number of drafts across datasets.}\label{fig:ab_num}
\myvspace{-10pt}
\crvspace{-10pt}
\end{figure} \subsection{Ablation Study I: Impact of Temperature}
We study the impact of different temperatures on the acceptance rates. The temperature affects the distributions of the target model and the draft model, even if the logits remain unchanged. It also affects the output text during the sampling process, resulting in different responses. \Cref{fig:ab_temperature} shows the results. 
We use LLaMA-7B as the target model and LLaMA-68M as the draft
model for our ablation studies.
We can have the following observations:
\begin{itemize}[leftmargin=*,noitemsep=0mm,topsep=-2pt]
    \item 
The impact of temperature is non-monotonic. Moreover, different methods respond differently to temperature changes.
\item
    At low temperatures, all methods fall into two categories. The first includes methods that allow duplicate tokens. When $T=0$, these methods essentially have only one effective draft token, the one with the largest logits on the draft model. The second includes methods that prevent duplicate tokens. When $T=0$, these methods always select the top $n$ tokens on the draft model.
    \item 
For sampling with replacement, the gap between the optimal acceptance rate and acceptance rates for previously existing verification methods, RRS and K-SEQ, gradually increases as the temperature rises.
\item 
As temperature increases, the gap between methods with replacement sampling and methods without replacement sampling decreases.
We can attribute this to the fact that,
    at high temperatures, the probability distribution is less concentrated, making with replacement sampling strategies have less probability to generate duplicate tokens. 
\end{itemize}

\begin{table}[t!]
\small
\centering
\setlength{\tabcolsep}{4pt}
\caption{
Acceptance rates of different MDSD methods on MT-Bench based on Eagle framework. 
}\label{tab:eagle}
\resizebox{1\linewidth}{!}{
\begin{tabular}{lcccccc}
\toprule
 \multirow{2}{*}{Method} & \multicolumn{2}{c}{\# Drafts = 2, \# Steps = 4} & \multicolumn{2}{c}{\# Drafts = 4, \# Steps = 3} & \multicolumn{2}{c}{
EAGLE default sparse tree
 }  \\
\cmidrule(r){2 - 3}\cmidrule(r){4 - 5} \cmidrule(r){6 - 7}
&   $\alpha$ & Speed  & $\alpha$ & Speed  & $\alpha$ & Speed   \\
\midrule
\multicolumn{7}{c}{$T=0.1$} \\
\midrule
  RRS w/ replacement & 75.3 $\pm$ 0.3 & -  & 78.4 $\pm$ 0.3& - & 74.7 $\pm$ 0.3 & -\\
 RRS w/o replacement & 79.4 $\pm$ 0.3 & 1.04 ($\pm$ 0.02) $\times$& 80.4 $\pm$ 0.3 &  1.03 ($\pm$ 0.01) $\times$  & 76.8 $\pm$ 0.3 & 1.04 ($\pm$ 0.02) $\times$\\
 SpecHub&  \textbf{84.0} $\pm$ 0.3 & 1.11 ($\pm$ 0.02) $\times$ & - & - & - & -\\
Greedy &  \textbf{84.7} $\pm$ 0.3 & 1.13 ($\pm$ 0.02) $\times$ &\textbf{88.8} $\pm$ 0.2 & 1.17 ($\pm$ 0.01) $\times$   & \textbf{79.1} $\pm$ 0.3 & 1.08 ($\pm$ 0.02) $\times$\\
  
\midrule
\multicolumn{7}{c}{$T=0.6$} \\
\midrule
  RRS w/ replacement & 78.8 $\pm$ 0.3 & - & 84.7 $\pm$ 0.3 & - & 76.2 $\pm$ 0.3 & -\\
 RRS w/o replacement & \textbf{82.4} $\pm$ 0.3 & 1.07 ($\pm$ 0.02) $\times$ & 88.6 $\pm$ 0.2 & 1.07 ($\pm$ 0.01) $\times$  & \textbf{77.6} $\pm$ 0.3 & 1.05 ($\pm$ 0.02) $\times$\\
 SpecHub& \textbf{82.3} $\pm$ 0.3 & 1.02 ($\pm$ 0.02) $\times$ & - & - & - & -\\
Greedy & \textbf{82.8} $\pm$ 0.3 & 1.04 ($\pm$ 0.02) $\times$ & \textbf{90.0} $\pm$ 0.2& 1.09 ($\pm$ 0.01) $\times$  & \textbf{78.3} $\pm$ 0.3 & 1.01 ($\pm$ 0.02) $\times$\\
  
\midrule
\multicolumn{7}{c}{$T=1.0$} \\
\midrule
  RRS w/ replacement & 76.7 $\pm$ 0.3& - & 83.5 $\pm$ 0.3 & - & 72.1 $\pm$ 0.3 & -\\
 RRS w/o replacement & 76.4 $\pm$ 0.3 & 1.00 ($\pm$ 0.02) $\times$  & 85.3 $\pm$ 0.3 & 1.05  ($\pm$ 0.01) $\times$ & \textbf{74.1} $\pm$ 0.3 & 1.03 ($\pm$ 0.02) $\times$\\
 SpecHub&  \textbf{79.5} $\pm$ 0.3 & 1.01 ($\pm$ 0.02) $\times$ &  - & - & - & -\\
Greedy & \textbf{79.2} $\pm$ 0.3 & 1.02 ($\pm$ 0.02) $\times$ & \textbf{87.8} $\pm$ 0.2 & 1.08 ($\pm$ 0.01) $\times$ & 72.9 $\pm$ 0.3 & 0.97 ($\pm$ 0.02) $\times$\\

\bottomrule
\end{tabular} }
\crvspace{-10pt}
\end{table} \myvspace{-2pt}
\subsection{Ablation Study II: Impact of Number of Drafts}
\myvspace{-2pt}
We investigate the impact of different numbers of drafts on the acceptance rates. The results are shown in \Cref{fig:ab_num}. We have the following observations: 
\begin{itemize}[leftmargin=*,noitemsep=0mm,topsep=-2pt]
    \item 
As the number of drafts increases, the coverage of the target model's possible outputs improves, therefore leading to better acceptance rate.
This trend holds for all methods.
    \item 
The draft sampling strategy significantly impacts the benefits derived from an increase in the number of drafts.
Sampling without replacement generally benefit more from an increase in drafts compared to sampling with replacement. This is because sampling with replacement can lead to redundant drafts, which do not fully leverage the advantages of increasing the number of drafts.
    \item 
For sampling with replacement, the gap between the optimal acceptance rate and acceptance rates for previously existing verification methods, RRS and K-SEQ, gradually increases as the number of drafts rises.

\end{itemize}

\crvspace{-5pt}
\subsection{Evaluating the Greedy Sampling Method on Generation Tasks}
\crvspace{-5pt}
In this section, we evaluate the effectiveness and generation efficiency of the proposed Greedy draft sampling method (\Cref{se:greedy_approach}) on real-world generation tasks and compare it with other MDSD methods.

We implement the Greedy method within the EAGLE Framework \citep{li2024eagle}, which supports multi-step MDSD with a draft tree structure. We experiment with three types of tree structures: (1) drafts = 2, depths = 4; (2) drafts = 4, depths = 3; and (3) a sparse tree with up to 4 drafts and 5 steps, which is the default setting in EAGLE. We conduct experiments on the MT-Bench dataset \mbox{\citep{zheng2023judging}} using Vicuna-7B-v1.3 \citep{vicuna2023} as the target model and its corresponding Eagle model with 0.24B parameters as the draft model.

\Cref{tab:eagle} presents the results. As discussed in \Cref{se:connection_to_spechub}, the Greedy method and SpecHub have equal acceptance rates when the number of draft tokens is 2. Our experiments confirm this theoretical insight, showing no statistically significant difference between the two methods for any temperature.

The Greedy method demonstrates improved performance at low temperatures. For example, at T=0.1, it achieves a higher acceptance rate compared to RRS without replacement, leading to faster generation. However, as the temperature increases, the performance gain of the Greedy method diminishes. This observation is consistent with the ablation study in \Cref{fig:ab_temperature}.

 \crvspace{-5pt}
\section{Conclusion}\label{se:conclusion}
\crvspace{-5pt}
In this paper, we studied the acceptance rate of Multi-Draft Speculative Decoding (MDSD). 

On the theoretical side, we discovered an equivalence between the optimal acceptance rate and a subset selection problem. We also provided efficient methods to compute the optimal acceptance rate for sampling with replacement and CPS.

On the practical side, for the first time, we measured the optimal acceptance rate under real text distributions and quantified the gap between existing algorithms and the optimal acceptance rate for sampling with replacement.

Furthermore, we proposed a practical greedy draft construction method that, in some cases, achieves an even higher acceptance rate than CPS.

We hope that our work will stimulate further research on improving the efficiency of large language model inference and make these powerful models more accessible and applicable in real-world scenarios. 
\subsubsection*{Acknowledgment}
This work was partially supported by NSF IIS 2347592, 2348169, DBI 2405416, CCF 2348306, CNS 2347617.

We are grateful to Jitendra Prajapati for identifying the conflation of successive sampling (SS) and conditional Poisson sampling (CPS) in our original acceptance calculation and providing an exact counterexample illustrating this distinction.

\bibliography{content/references}
\bibliographystyle{plainnat}

\newpage
\appendix
\onecolumn
\section{Approximate Solutions}\label{se:approxverify}

\subsection{Recursive Rejection Sampling (RRS)}
\citet{yang2024multi} and \citet{jeon2024recursive} use the Recursive Rejection Sampling method. Define the residual distribution $\opn{Res}^{p-q}\in \Delta_\Sigma$ for $p,q\in \Delta_\Sigma$ as:
\begin{equation}
\opn{Res}^{p-q}(i)=\frac{(p(i)-q(i))_+}{\sum_{z\in\Sigma}(p(z)-q(z))_+}
\end{equation}
When $p=q$, $\opn{Res}^{p-q}$ can be defined as an arbitrary distribution.

For $p_{\opn{draft}}$ from sampling with replacement, the RRS algorithm is recursively defined as:
\begin{equation}
\pi^{\opn{RRS},\opn{w}}_{p,p_{\opn{draft}}}(i|\bar{i})=\widetilde{\pi}^{\opn{RRS},\opn{w}}_{p,q}(i|\bar{i})
\end{equation}
where
\begin{equation}
\widetilde{\pi}^{\opn{RRS},\opn{w}}_{p,q}(i|\bar{i})=
\begin{cases}\min(\frac{p(\bar{i}_1)}{q(\bar{i}_1)},1)&i=\bar{i}_1\\(1-\frac{p(\bar{i}_1)}{q(\bar{i}_1)})_+
\widetilde{\pi}^{\opn{RRS},\opn{w}}_{\opn{Res}^{p-q},q}(i|\bar{i}_{2:})
&i\neq \bar{i}_1\end{cases}
\end{equation}
and
\begin{equation}
\widetilde{\pi}^{\opn{RRS},\opn{w}}_{p,q}(i|())=p(i)
\end{equation}
Here $\bar{i}_{2:}$ denotes the sequence $\bar{i}$ with the first element removed.

For $p_{\opn{draft}}$ from successive sampling in \eqref{eq:successive}, the RRS algorithm is defined as: These proposal updates are specific to SS and do not give the conditional proposals for CPS.
\begin{equation}
\pi^{\opn{RRS},\opn{w}}_{p,p_{\opn{draft}}}(i|\bar{i})=\widetilde{\pi}^{\opn{RRS},\opn{wo}}_{p,q}(i|\bar{i})
\end{equation}
where
\begin{equation}
\widetilde{\pi}^{\opn{RRS},\opn{wo}}_{p,q}(i|\bar{i})=
\begin{cases}\min(\frac{p(\bar{i}_1)}{q(\bar{i}_1)},1)&i=\bar{i}_1\\(1-\frac{p(\bar{i}_1)}{q(\bar{i}_1)})_+
\widetilde{\pi}^{\opn{RRS},\opn{wo}}_{\opn{Res}^{p-q},q^{\neg \bar{i}_1}}(i|\bar{i}_{2:})
&i\neq \bar{i}_1\end{cases}
\end{equation}
and
\begin{equation}
\widetilde{\pi}^{\opn{RRS},\opn{wo}}_{p,q}(i|())=p(i)
\end{equation}

The acceptance rates are denoted as $\alpha^{\opn{RRS},\opn{w}}(p,p_{\opn{draft}})$ and $\alpha^{\opn{RRS},\opn{wo}}(p,p_{\opn{draft}})$, respectively.

\subsection{K-SEQ}\label{se:kseq}
\citet{sun2024spectr} proposed the K-SEQ method to verify drafts sampled with replacement. Define
\begin{equation}
\beta_{p,q}(\rho)=\sum_{i\in\Sigma}\min(\frac{p(i)}{\rho},q(i))
\end{equation}
and let $\rho$ be the solution to the equation
\begin{equation}
1-(1-\beta_{p,q}(\rho))^n=\rho\beta_{p,q}(\rho)
\end{equation}

The K-SEQ algorithm is defined as:
\begin{equation}
\pi^{\opn{K-SEQ}}_{p,p_{\opn{draft}}}(i|\bar{i})=\widetilde{\pi}^{\opn{K-SEQ}}_{p,q,\rho}(i|\bar{i})
\end{equation}
where
\begin{equation}
\widetilde{\pi}^{\opn{K-SEQ}}_{p,q,\rho}(i|\bar{i})=
(1-\frac{p(\bar{i}_1)}{\rho q(\bar{i}_1)})_+
\widetilde{\pi}^{\opn{K-SEQ}}_{p,q,\rho}(i|\bar{i}_{2:})
+
\begin{cases}\min(\frac{p(\bar{i}_1)}{\rho q(\bar{i}_1)},1)&i=\bar{i}_1\\0&i\neq \bar{i}_1\end{cases}
\end{equation}
and
\begin{equation}
\widetilde{\pi}^{\opn{K-SEQ}}_{p,q,\rho}(i|())=\frac{p(i)-\min\left\{q(i),\frac{p(i)}\rho\right\}\frac{1-(1-\beta_{p,q}(\rho))^n}{\beta_{p,q}(\rho)}}{(1-\beta_{p,q}(\rho))^n}
\end{equation}

The acceptance rate is denoted as $\alpha^{\opn{K-SEQ}}(p,p_{\opn{draft}})=1-(1-\beta_{p,q}(\rho))^n$, which is theoretically guaranteed to achieve a $(1-e^{-1})$-approximation of the optimal acceptance rate. \section{Related Works}\label{se:related_work}
\subsection{Draft Model Design}
Numerous studies have explored the design of better draft models for speculative decoding. In principle, any autoregressive probabilistic model can serve as a draft model. The simplest approaches include using n-gram models \citep{ou2024lossless} or document retrieval as draft models \citep{yang2023inference,he2023rest}. Small transformer-based language models have also been employed \citep{leviathan2023fast,chen2023accelerating}, often with distillation techniques to further increase the overlap between the draft and target models \citep{zhou2023distillspec}.

The design of a good draft model involves a trade-off between its similarity to the target model and its computational complexity. More complex draft models lead to higher acceptance rates due to their closer resemblance to the target model, but they also incur higher computational overhead. To achieve a better trade-off, some works have proposed reusing the target model's computational results. For example, \citet{monea2023pass} use the original model with ``look ahead" tokens, while \citet{cai2024medusa} add new heads to the last hidden layer of the original model to predict tokens further ahead. \citet{li2024eagle} reuse the last layer hidden state computation of the large model and introduce a new attention layer to predict the next token. \citet{sun2024triforce} employ the target model with a partial key-value cache as the draft model.

\subsection{Multi-Draft Speculative Decoding}\label{sec:related_mdsd}
Many related works on Multi-Draft Speculative Decoding (MDSD) have been introduced in other sections. This paper focuses on the single-step Multi-Draft scenario. When MDSD generates multiple steps, with each step involving multiple drafts, it forms a tree structure. Sequoia \citep{chen2024sequoia} propose a dynamic programming algorithm to search for the optimal tree topology.

As the tree grows deeper, the acceptance probability of certain branches decreases. Cascade Speculative Drafting \citep{chen2023cascade} addresses this issue by assigning the largest draft model to generate draft tokens at shallower levels, which are more likely to be accepted, and gradually using smaller models to generate drafts for less relevant branches.

\citet{khisti2024importanceweighted} studied the optimal acceptance rate for special case of sampling with replacement for $n=2$ drafts, and obtained the following result:
\begin{equation}
\alpha^\ast(p,p_{\opn{draft}}) = \min_{H \subset\Sigma} \left\{
\sum_{i\in H} p(i)+\left(\sum_{i\in \Sigma\setminus H} q(s)\right)^2+2\left(\sum_{i\in H} q(s)\right)\left(\sum_{i\in \Sigma\setminus H} q(s)\right)
\right\}
~.
\end{equation}

This is essentially the same as our result \eqref{eq:subset} under this special case. 
However, our theory is more general, without any assumption on the draft sampling methods or the number of draft tokens.

\subsection{Multi-Step Speculative Decoding}\label{sec:related_multi_step}
The basic single-step, single-draft speculative decoding, as introduced in \Cref{sec:prelim_basic_sps}, can be applied to multiple steps, with each step having only one draft and an independent verification process \citep{leviathan2023fast,chen2023accelerating}. However, such an approach of repeatedly applying single-step verification is not optimal for the multi-step scenario. Some works, such as \citet{sun2024optimal,huaccelerated,sunblock}, have designed better verification algorithms specifically for the multi-step setting. These algorithms are tailored for the multi-step scenario while remaining compatible with the single-step case, reducing to the basic speculative sampling algorithm when applied to a draft sequence of length 1.

\begin{figure}[h]
\centering
\begin{tikzpicture}[
    box/.style={draw, rectangle, minimum width=4.5cm, minimum height=1.2cm, align=center},
    smallbox/.style={draw, rectangle, minimum width=3.5cm, minimum height=1.2cm, align=center},
    arrow/.style={->, thick},
]

\node[box] (base) at (0,0) {Speculative Decoding \\ \citet{leviathan2023fast} \\ \citet{chen2023accelerating}};

\node[smallbox] (ms1) at (6,0) {\citet{sun2024optimal} \\ \citet{huaccelerated} \\ \citet{sunblock}};

\node[box] (md1) at (0,-3) {\citet{sun2024spectr} \\ \citet{khisti2024importanceweighted} \\ This paper};

\draw[arrow] (base) -- node[above] {multi-step} (ms1);
\draw[arrow] (base) -- node[left] {multi-draft} (md1);

\end{tikzpicture}

\caption{Different directions for improving speculative decoding.}
\label{fig:related_work}
\end{figure}
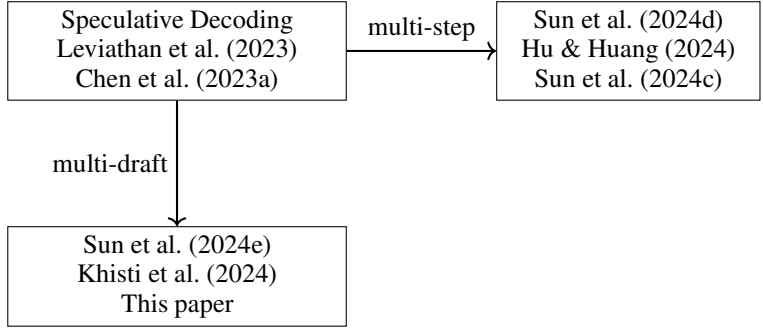

Multi-step speculative decoding and multi-draft speculative decoding represent different directions for improvement.

As shown in \Cref{fig:related_work}, \citet{sun2024spectr,khisti2024importanceweighted} and our work improve speculative decoding from the multi-draft perspective. When there is only a single draft, it reduces to the case in \citet{leviathan2023fast,chen2023accelerating}. On the other hand, \citet{sun2024optimal,huaccelerated,sunblock} enhance speculative decoding from the multi-step perspective. When there is only a single step, it reduces to the case in \citet{leviathan2023fast,chen2023accelerating}.

Combining both improvements in the multi-draft and multi-step scenario would be ideal, and could be a direction for future research. \section{Proofs}\label{se:proofs}

\begin{proof}[Proof of \Cref{le:eq_lp_lp_prime}]\label{pr:eq_lp_lp_prime}
Let $f_1(C)$ and $f_2(S)$ denote the objective function values of \eqref{eq:lp} and \eqref{eq:lp_prime}, respectively. Let $v_1=f_1(C^\ast)$ and $v_2=f_2(S^\ast)$ be the optimal objective function values.

First, we show that the optimal solution of \eqref{eq:lp} is feasible for \eqref{eq:lp_prime}. Define $S_{i,\bar{i}}=C^\ast_{i,\bar{i}}$ for $\bar{i}\in A_i$ and $S_{i,\bar{i}}=0$ for $\bar{i}\notin A_i$. This solution maintains the objective function value, i.e., $f_2(S)=f_1(C^\ast)$. Therefore, $v_2=f_2(S^\ast)\geq f_2(S)=f_1(C^\ast)=v_1$.

Next, we show that the optimal solution of \eqref{eq:lp_prime} is feasible for \eqref{eq:lp}. Define $p^{\opn{res}}(i)=p(i)-\sum_{\bar{i}\in\Sigma^n}S^\ast_{i,\bar{i}}\geq 0$ for $i\in\Sigma$ and $p_{\opn{draft}}^{\opn{res}}(\bar{i})=p_{\opn{draft}}(\bar{i})-\sum_{i\in\Sigma}S^\ast_{i,\bar{i}}\geq 0$ for $\bar{i}\in\Sigma^n$. We have $\sum_{\bar{i}\in\Sigma^n}p_{\opn{draft}}^{\opn{res}}(\bar{i})=\sum_{i\in\Sigma}p^{\opn{res}}(i)$. Define $C_{i,\bar{i}}=S^\ast_{i,\bar{i}}+\frac{p^{\opn{res}}(i)p_{\opn{draft}}^{\opn{res}}(\bar{i})}{\sum_{i\in\Sigma}p^{\opn{res}}(i)}\geq S^\ast_{i,\bar{i}}$. This solution has a larger objective function value, i.e., $f_1(C)\geq f_1(S^\ast)$. Therefore, $v_1=f_1(C^\ast)\geq f_1(C)\geq f_2(S^\ast)=v_2$.

Combining the two parts, we have $v_1=v_2$, which proves the equivalence of the two formulations.
\end{proof} \begin{proof}[Proof of \Cref{th:q_convex_wr}]
For $\bar{i}=(\bar{i}_1,\dots,\bar{i}_n)\in\Sigma^n$, we have $p_{\opn{draft}}(\bar{i})=\prod_{j=1}^n q(\bar{i}_j)$. The function $Q(H)=\sum_{\bar{i}\in H^n} p_{\opn{draft}}(\bar{i})$ represents the probability that all $n$ samples drawn with replacement are in the set $H$. Therefore, writing $s=\sum_{i\in H}q(i)$, we have $Q(H)=s^n$.

Consider the convex function $g(t)=t^n$ on $[0,\infty)$. To prove the $q$-convexity of $Q$, it suffices to show that for all $H\subset \Sigma$ and $x,y\in \Sigma\setminus H$ with $x\neq y$ and $q(x),q(y)>0$, we have:
\begin{equation}
\frac{(s+q(x))^n-s^n}{q(x)}\leq\frac{(s+q(x)+q(y))^n-(s+q(x))^n}{q(y)}
\end{equation}

This can be rewritten as:
\begin{equation}
\frac{g(s+q(x))-g(s)}{q(x)}\leq\frac{g(s+q(x)+q(y))-g(s+q(x))}{q(y)}
\end{equation}

Note that both sides are finite differences of the convex function $g$. Define $a=s$, $b=s+q(x)$, and $c=s+q(x)+q(y)$. It suffices to show that:
\begin{equation}
\frac{g(b)-g(a)}{b-a}\leq\frac{g(c)-g(b)}{c-b}
\end{equation}

This follows directly from the convexity of $g$. Zero-probability tokens can be omitted.
\end{proof} \begin{proof}[Proof of \Cref{th:q_convex_wo}]
Work on $\operatorname{supp}(q)$. Let $T_i\sim\operatorname{Exp}(q(i))$ independently, and let $S$ contain the first $n$ arrivals. Exponential races select each next token in proportion to its remaining weight. Memorylessness therefore gives exactly \eqref{eq:successive}.
For $n=1$, $Q(H)=\sum_{i\in H}q(i)$ and the claim is immediate. Assume $n\geq2$ and fix distinct $x,y\notin H$. If $|H|<n-1$, then $Q(x|H)=0$, so the required inequality is trivial.

Otherwise, condition on the clocks $T_H$. Let $u$ and $v$ be their $(n-1)$st and $n$th arrival times, taking $v=\infty$ if $|H|=n-1$. Write $a=q(x)$, $b=q(y)$, $c=\sum_{i\notin H\cup\{x,y\}}q(i)$, and $J=\int_u^v e^{-(a+b+c)t}\,dt$. For the marginal events $A=\{S\subseteq H\cup\{x\},\ x\in S\}$ and $B=\{S\subseteq H\cup\{x,y\},\ y\in S\}$,
\begin{align*}
\frac{\Pr(A\mid T_H)}{a}
&=e^{-(b+c)u}\int_0^u e^{-at}\,dt+J,\\
\frac{\Pr(B\mid T_H)}{b}
&\geq e^{-cu}\int_0^u e^{-bt}\,dt+J.
\end{align*}
The first line splits according to $T_x<u$ or $u\leq T_x<v$. For the second, sufficient events are: $T_y<u$ with all clocks outside $H\cup\{x,y\}$ exceeding $u$, or $u\leq T_y<v$ with $T_x$ and all those outside clocks exceeding $T_y$. These disjoint events give the two terms of the lower bound.
For $0\leq t\leq u$, $e^{-bt}\geq e^{-bu}e^{-at}$, so the second line is at least the first. Averaging over $T_H$ yields $Q(x|H)/q(x)\leq Q(y|H\cup\{x\})/q(y)$.
\end{proof}
 \begin{proof}[Proof of \Cref{th:q_convex_cb}]
By \eqref{eq:cb_law}, $Q(H)=W_{n,H}/W_{n,\Sigma}$, with $W_{n,\Sigma}>0$.
To prove the $q$-convexity of $Q$, it suffices to show the defining inequality for distinct $x,y\notin H$ with positive weights. The recurrence $W_{n,H\cup\{x\}}=W_{n,H}+q(x)W_{n-1,H}$ gives
\[
\frac{Q(x|H)}{q(x)}
=\frac{W_{n-1,H}}{W_{n,\Sigma}}
\leq\frac{W_{n-1,H\cup\{x\}}}{W_{n,\Sigma}}
=\frac{Q(y|H\cup\{x\})}{q(y)}.
\]
The inequality holds because all weights are nonnegative. The convention $W_{0,H}=1$ also covers $n=1$.
\end{proof}
 \begin{proof}[Proof of \Cref{th:q_convex_supermodular}]
It suffices to show that for all $H\subset \Sigma$ and $x,y\in \Sigma\setminus H$ with $x\neq y$, we have:
\begin{equation}
Q(x|H)\leq Q(x|H\cup\{y\})
\end{equation}

By the $q$-convexity of $Q$, we have:
\begin{equation}
\frac{Q(x|H)}{q(x)}\leq\frac{Q(y|H\cup\{x\})}{q(y)}
\end{equation}

Therefore,
\begin{equation}
\frac{Q(x|H)}{q(x)}\leq\frac{Q(x|H)+Q(y|H\cup\{x\})}{q(x)+q(y)}
=\frac{Q(H\cup\{x,y\})-Q(H)}{q(x)+q(y)}
\leq\frac{Q(y|H\cup\{x\})}{q(y)}
\end{equation}

Similarly, by symmetry, we can reverse $x$ and $y$ to obtain:
\begin{equation}
\frac{Q(y|H)}{q(y)}\leq\frac{Q(H\cup\{x,y\})-Q(H)}{q(x)+q(y)}\leq\frac{Q(x|H\cup\{y\})}{q(x)}
\end{equation}

Therefore,
\begin{equation}
\frac{Q(x|H)}{q(x)}\leq\frac{Q(H\cup\{x,y\})-Q(H)}{q(x)+q(y)}\leq\frac{Q(x|H\cup\{y\})}{q(x)}
\end{equation}

This implies that:
\begin{equation}
Q(x|H)\leq Q(x|H\cup\{y\})
\end{equation}
\end{proof} \begin{proof}[Proof of \Cref{th:efficient_computation}]
We have:
\begin{align}
f(x|H)&=p(x)-Q(x|H)\\
&=p(x)(1-\frac{q(x)}{p(x)}\frac{Q(x|H)}{q(x)})\leq0
\end{align}

Therefore,
\begin{equation}
\frac{f(x|H)}{p(x)}=1-\frac{q(x)}{p(x)}\frac{Q(x|H)}{q(x)}\leq0
\end{equation}

By assumption, $\frac{q(x)}{p(x)}\leq\frac{q(y)}{p(y)}$. By the $q$-convexity of $Q$, we have $\frac{Q(x|H)}{q(x)}\leq\frac{Q(y|H\cup\{x\})}{q(y)}$. Therefore,
\begin{equation}
\frac{q(y)}{p(y)}\frac{Q(y|H\cup\{x\})}{q(y)}\geq\frac{q(x)}{p(x)}\frac{Q(x|H)}{q(x)}
\end{equation}

It follows that:
\begin{align}
\frac{f(y|H\cup\{x\})}{p(y)}&=1-\frac{q(y)}{p(y)}\frac{Q(y|H\cup\{x\})}{q(y)}\\
&\leq\frac{f(x|H)}{p(x)}\leq0
\end{align}
\end{proof} \begin{proof}[Proof of \Cref{th:greedy_optimal}]
We first prove that the acceptance rate of the greedy method is:
\begin{align}
&\alpha^{\opn{Greedy}}(p,p_{\opn{draft}})
\\=&
\sum_{i\in\Sigma} \sum_{\bar{i}\in A_i}\pi^{\opn{Greedy}}_{p,p_{\opn{draft}}}(i,\bar{i})
\\=&
\sum_{i\in\Sigma} \sum_{\bar{i}\in A_i}\pi^{\opn{Greedy}}_{p,p_{\opn{draft}}}(i|\bar{i}) p_{\opn{draft}}(\bar{i})
\\=&
\sum_{i\in\opn{Top}_{n-1}(q)} \sum_{\bar{i}\in A_i}\pi^{\opn{Greedy}}_{p,p_{\opn{draft}}}(i|\bar{i}) p_{\opn{draft}}(\bar{i})
\\&+
\sum_{i\in\Sigma\setminus\opn{Top}_{n-1}(q)} \sum_{\bar{i}\in A_i}\pi^{\opn{Greedy}}_{p,p_{\opn{draft}}}(i|\bar{i}) p_{\opn{draft}}(\bar{i})
\\=&
\sum_{i\in\opn{Top}_{n-1}(q)} \sum_{\bar{i}_n\in \Sigma}\pi^\ast_{p,q^{\neg \opn{Top}_{n-1}(q)}}(i|\bar{i}_n) q^{\neg \opn{Top}_{n-1}(q)}(\bar{i}_n)
\\&+
\sum_{i\in\Sigma\setminus\opn{Top}_{n-1}(q)} \pi^{\opn{Greedy}}_{p,p_{\opn{draft}}}(i|(\opn{Top}_{n-1}(q),i)) q^{\neg \opn{Top}_{n-1}(q)}(i)
\\=&
\sum_{i\in\opn{Top}_{n-1}(q)} p(i)
\\&+
\sum_{i\in\Sigma\setminus\opn{Top}_{n-1}(q)} \pi^\ast_{p,q^{\neg \opn{Top}_{n-1}(q)}}(i|i) q^{\neg \opn{Top}_{n-1}(q)}(i)
\\=&
\sum_{i\in\opn{Top}_{n-1}(q)} p(i)
+
\sum_{i\in\Sigma\setminus\opn{Top}_{n-1}(q)} \min(p(i),q^{\neg \opn{Top}_{n-1}(q)}(i))
\end{align}

Note that $\sum_{i\in\opn{Top}_{n-1}(q)} \min(p(i),q^{\neg \opn{Top}_{n-1}(q)}(i))=\sum_{i\in\opn{Top}_{n-1}(q)} \min(p(i),0)=0$.

Next, we compute the optimal acceptance rate. Note that when $\opn{Top}_{n-1}(q)\nsubseteq H$, we must have $Q(H)=0$. When $\opn{Top}_{n-1}(q)\subseteq H$, we have $Q(H)=\sum_{i\in H} q^{\neg \opn{Top}_{n-1}(q)}(i)$. Therefore,
\begin{align}
&\alpha^\ast(p,p_{\opn{draft}})
\\=&
1+\min_{H\subset\Sigma} P(H)-Q(H)
\\=&
1+\min_{H\subset\Sigma,\text{s.t.}\opn{Top}_{n-1}(q)\subseteq H} \sum_{i\in H} p(i)-q^{\neg \opn{Top}_{n-1}(q)}(i)
\end{align}

The optimal set is $H^\ast=\{i\in\Sigma|q^{\neg \opn{Top}_{n-1}(q)}(i)\geq p(i)\}\cup\opn{Top}_{n-1}(q)$. In this case,
\begin{align}
&\alpha^\ast(p,p_{\opn{draft}})
\\=&
1-\sum_{i\in\Sigma}(q^{\neg \opn{Top}_{n-1}(q)}(i)-p(i))_+ + \sum_{i\in\opn{Top}_{n-1}(q)} p(i)
\\=&
\sum_{i\in\Sigma}\min(p(i),q^{\neg \opn{Top}_{n-1}(q)}(i))_+ + \sum_{i\in\opn{Top}_{n-1}(q)} p(i)
\end{align}
\end{proof} \subsection{Derivation of the Dual Problem}\label{se:dual}
We start from the primal problem \eqref{eq:lp_prime}:
\begin{equation}
\begin{aligned}
\max_{S\in\mathbb{R}^{\Sigma\times\Sigma^n}} & \sum_{i\in\Sigma} \sum_{\bar{i}\in \Sigma^n} S_{i,\bar{i}} &
\\
\text{s.t.} & \sum_{\bar{i}\in\Sigma^n}S_{i,\bar{i}}\leq p(i)& \forall i\in\Sigma
\\
& \sum_{i\in\Sigma}S_{i,\bar{i}}\leq {p_{\opn{draft}}}(\bar{i})& \forall \bar{i}\in\Sigma^n
\\
& S_{i,\bar{i}}\geq0 & \forall i\in\Sigma,\bar{i}\in\Sigma^n
\\
& S_{i,\bar{i}}=0 & \forall i\in\Sigma,\bar{i}\notin A_i
\end{aligned}
\end{equation}

We introduce dual variables $y_i$ for each constraint $\sum_{\bar{i}\in\Sigma^n}S_{i,\bar{i}}\leq p(i)$ and $z_{\bar{i}}$ for each constraint $\sum_{i\in\Sigma}S_{i,\bar{i}}\leq {p_{\opn{draft}}}(\bar{i})$. The Lagrangian function is:
\begin{align}
L(S,y,z)=&\sum_{i\in\Sigma} \sum_{\bar{i}\in \Sigma^n} S_{i,\bar{i}}\\
&+\sum_{i\in\Sigma} y_i (p(i)-\sum_{\bar{i}\in\Sigma^n}S_{i,\bar{i}})\\
&+\sum_{\bar{i}\in\Sigma^n} z_{\bar{i}} (p_{\opn{draft}}(\bar{i})-\sum_{i\in\Sigma}S_{i,\bar{i}})
\end{align}

The dual function is:
\begin{equation}
\begin{aligned}
    g(y,z) = 
    \max_{S \in \mathbb{R}^{\Sigma \times \Sigma^n}} & L(S,y,z) & \\
\text{s.t.~~} & S_{i,\bar{i}} \geq 0 & \forall i \in \Sigma, \bar{i} \in \Sigma^n \\ 
& S_{i,\bar{i}} = 0 & \forall i \in \Sigma, \bar{i} \notin A_i
\end{aligned}
\end{equation}

Rearranging the Lagrangian function:
\begin{align}
L(S,y,z)=&\sum_{i\in\Sigma} \sum_{\bar{i}\in \Sigma^n} (1-y_i-z_{\bar{i}}) S_{i,\bar{i}}\\
&+\sum_{i\in\Sigma} y_i p(i)+\sum_{\bar{i}\in\Sigma^n} z_{\bar{i}} p_{\opn{draft}}(\bar{i})
\end{align}

For the dual function to be bounded, we must have:
\begin{equation}
1-y_i-z_{\bar{i}}\leq0, \forall i\in\Sigma,\bar{i}\in A_i
\end{equation}

Therefore, the dual problem is:
\begin{equation}
\begin{aligned}
\min_{y\in\mathbb{R}^{\Sigma},z\in\mathbb{R}^{\Sigma^n}} & \sum_{i\in\Sigma} y_i p(i)+ \sum_{\bar{i}\in\Sigma^n} z_{\bar{i}} p_{\opn{draft}}(\bar{i}) &
\\
\text{s.t.} & y_i+z_{\bar{i}}\geq1 & \forall i\in\Sigma,\bar{i}\in A_i
\\
& y_i\geq 0& \forall i\in\Sigma
\\
& z_{\bar{i}}\geq 0& \forall \bar{i}\in\Sigma^n
\end{aligned}
\end{equation}

This completes the derivation of the dual problem.

\section{Additional Illustration}\label{sec:add_ill}
Illustration of single-step draft tokens generation and verification:

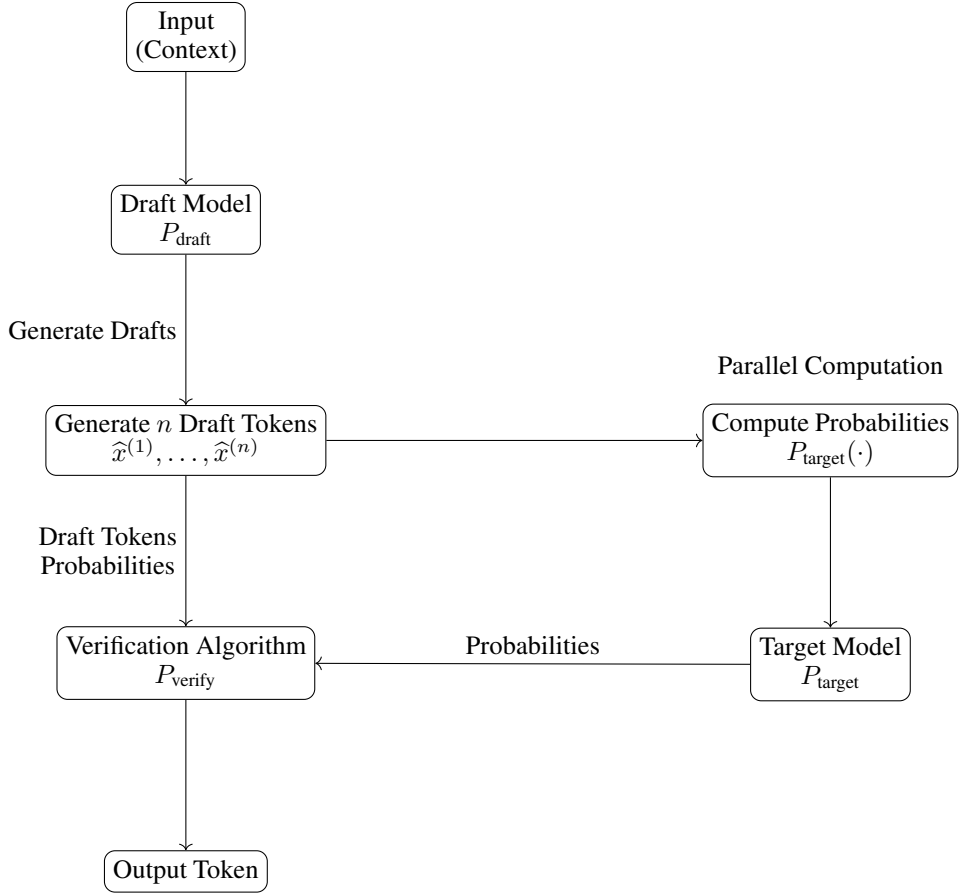
\begin{tikzpicture}[
    node distance=1.5cm,
    box/.style={
draw=black, 
    rounded corners, align=center},
]

\node[box] (start) {Input\\(Context)};
\node[box, below=of start] (draftModel) {Draft Model\\$P_{\text{draft}}$};
\node[box, below=2cm of draftModel] (generateDrafts) {Generate $n$ Draft Tokens\\$\widehat{x}^{(1)}, \dots, \widehat{x}^{(n)}$};
\node[box, below=2cm of generateDrafts] (verification) {Verification Algorithm\\$P_{\text{verify}}$};
\node[box, below=2cm of verification] (output) {Output Token};

\node[box, right=5cm of generateDrafts] (computeProbs) {Compute Probabilities\\$P_{\text{target}}(\cdot)$};
\node[box, below=2cm of computeProbs] (targetModel) {Target Model\\$P_{\text{target}}$};

\draw[->] (start) -- (draftModel);
\draw[->] (draftModel) -- node[left] {Generate Drafts} (generateDrafts);
\draw[->] (generateDrafts) -- node[left, align=center] {Draft Tokens\\Probabilities} (verification);
\draw[->] (verification) -- node[left] {} (output);

\draw[->] (generateDrafts) -- (computeProbs);
\draw[->] (computeProbs) -- (targetModel);
\draw[->] (targetModel) -- node[above] {Probabilities} (verification);

\node[above=0.2cm of computeProbs] {Parallel Computation};

\end{tikzpicture}

Pseudo code for apply multi-draft speculative sampling for multiple steps, with arbitrary tree topology.
\begin{lstlisting}[language=Python]
def multi_draft_speculative_decoding(prompt,
                                     tree_topology, 
                                     draft_model, 
                                     target_model):
    """
    Multi-Draft Speculative Decoding algorithm for accelerating
    language model inference.
   
    Example tree_topology:
        tree_topology = [
            [0], [1], [2], [3],  # First level: 4 branches
            [0,0], [0,1], [0,2], [1,0], [1,1], 
            [2,0], [2,1], [3,0],  # Second level
            [0,0,0], [0,0,1], [0,0,2], [0,1,0], 
            [0,1,1], [0,2,0], [0,2,1], [1,0,0], # Third level
            [0,0,0,0], [0,0,0,1], [0,0,0,2],  # Fourth level
            [0,0,0,0,0], [0,0,0,0,1]  # Fifth level
        ]
        This is the default EAGLE tree structure where each number 
        represents which draft token to use at each level.
    """
    # Initialize dictionaries to store drafts and distributions
    drafts = {}
    draft_distributions = {}
    target_distributions = {}
   
    # Generate drafts for each prefix in the tree topology
    for prefix in [[]] + tree_topology:
        children = get_children(tree_topology, prefix)
        if not children:
            continue  # Skip if no expandable children paths
        # Generate drafts and corresponding distributions
        # e.g. sampling with/without replacement or Section 5.1
        (
            drafts[tuple(prefix)], 
            draft_distributions[tuple(prefix)]
        ) = generate_draft_tokens(draft_model, prefix)
   
    # Compute probability distributions from target model for all drafts
    target_distributions = compute_target_distributions(target_model, drafts)
   
    # Start verification and generation process
    prefix = []
    while True:
        if tuple(prefix) not in drafts:
            break  # End if no available drafts
        # e.g. RRS or K-Seq or Section 5.2
        output_token = verification(
            drafts[tuple(prefix)],
            draft_distributions[tuple(prefix)],
            target_distributions[tuple(prefix)]
        )
        prefix.append(output_token)
   
    # Return the generated sequence
    return prefix

def get_children(tree_topology, prefix):
    """
    Get all child paths in tree_topology that extend the given
    prefix by one token.
   
    Examples:
        tree_topology = [[0], [1], [0,0], [0,1], [1,0]]
        get_children([], tree_topology) -> [[0], [1]]
        get_children([0], tree_topology) -> [[0,0], [0,1]]
        get_children([1], tree_topology) -> [[1,0]]
        get_children([0,0], tree_topology) -> []
    """
    return [
        path for path in tree_topology
        if len(path) == len(prefix) + 1 and path[:len(prefix)] == prefix
    ]
\end{lstlisting}

Unfolded definition of verify algorithm for greedy draft construction.
\begin{equation}
\begin{aligned}
&\pi^{\opn{Greedy}}_{p,p_{\opn{draft}}}(i|\bar{i})
\\=&\pi^\ast_{p,q^{\neg \opn{Top}_{n-1}(q)}}(i|\bar{i}_n)
\\=&\begin{cases}
\min(\frac{p(i)(1-\sum_{j\in\opn{Top}_{n-1}(q)} q(j))}{q(i)},1)&i=\bar{i}_n
\\
(1-\frac{p(\bar{i}_n)(1-\sum_{j\in\opn{Top}_{n-1}(q)} q(j))}{q(\bar{i}_n)})_+
\frac{p(i)(1-\sum_{j\in\opn{Top}_{n-1}(q)} q(j))}{\sum_{z\in\Sigma} (p(z)(1-\sum_{j\in\opn{Top}_{n-1}(q)} q(j))-\mathbb{I}(z\notin\opn{Top}_{n-1}(q))q(z))_+}&i\in \opn{Top}_{n-1}(q)
\\
(1-\frac{p(\bar{i}_n)(1-\sum_{j\in\opn{Top}_{n-1}(q)} q(j))}{q(\bar{i}_n)})_+
\frac{(p(i)(1-\sum_{j\in\opn{Top}_{n-1}(q)} q(j))-q(i))_+}{\sum_{z\in\Sigma} (p(z)(1-\sum_{j\in\opn{Top}_{n-1}(q)} q(j))-\mathbb{I}(z\notin\opn{Top}_{n-1}(q))q(z))_+}
&i\neq\bar{i}_n, i\notin \opn{Top}_{n-1}(q)
\end{cases}
\end{aligned}
\end{equation}

\section{Summary of Notations}
\begin{itemize}
\item $\Sigma$: The vocabulary set
\item $\Delta_\Sigma$: The probability simplex over vocabulary $\Sigma$
\item $[n]$: The set {1,...,n}
\item $P_{\opn{target}}(\cdot|x_1,x_2,...,x_m)$: The target model, a probabilistic model that predicts the probability of the next word given the context
\item $P_{\opn{draft}}(\cdot|x_1,x_2,...,x_m)$: The draft model used to generate candidate tokens
\item $P_{\opn{verify}}(\cdot|\widehat{x}^{(1)},\dots\widehat{x}^{(n)})$: The verification algorithm that selects the final output token from the draft tokens
\item $p(\cdot)=P_{\opn{target}}(\cdot|x_1,x_2,...,x_m)$: Shorthand for the target distribution
\item $p_{\opn{draft}}$: The distribution of draft tokens
\item $\pi\in\Pi(p,p_{\opn{draft}})$: A joint distribution with marginal distributions $p$ and $p_{\opn{draft}}$
\item $\pi^\ast_{p,p_{\opn{draft}}}\in\Pi(p,p_{\opn{draft}})$: The optimal transport joint distribution
\item $\alpha^\ast(p,p_{\opn{draft}})$: The optimal acceptance rate
\item $A_i:=\{\bar{i}\in \Sigma^n|\exists j\in[n], \overline{i}_j=i\}$: The incidence set for token $i$
\item $q(\cdot)$: The shorthand notation of the output distribution of the draft model
\item $q^{\neg \bar{i}_1,\dots,\bar{i}_{j-1}}(x)$: The probability of token $x$ when sampling without replacement, excluding previously selected tokens
\item $\opn{Res}^{p-q}\in \Delta_\Sigma$: The residual distribution
\item $\pi^{\opn{RRS},\opn{w}}_{p,p_{\opn{draft}}}$, $\pi^{\opn{RRS},\opn{wo}}_{p,p_{\opn{draft}}}$: The RRS verification algorithms for with/without replacement sampling
\item $\alpha^{\opn{RRS},\opn{w}}(p,p_{\opn{draft}})$, $\alpha^{\opn{RRS},\opn{wo}}(p,p_{\opn{draft}})$: Acceptance rates for RRS with/without replacement
\item $\beta_{p,q}(\rho)$: A function used in the K-SEQ algorithm
\item $\pi^{\opn{K-SEQ}}_{p,p_{\opn{draft}}}$: The K-SEQ verification algorithm
\item $\alpha^{\opn{K-SEQ}}(p,p_{\opn{draft}})$: Acceptance rate for the K-SEQ algorithm
\item $P(H)=\sum_{i\in H} p(i)$: Sum of target probabilities over set $H$
\item $Q(H)=\sum_{\bar{i}\in H^n} p_{\opn{draft}}(\bar{i})$: Sum of draft probabilities over set $H^n$
\item $f(H)=P(H)-Q(H)$: Difference between target and draft probabilities over set $H$
\item $\pi^{\opn{Greedy}}_{p,p_{\opn{draft}}}$: The greedy verification algorithm
\item $\alpha^{\opn{Greedy}}(p,p_{\opn{draft}})$: Acceptance rate for the greedy draft sampling method
\item $C_{i,j}$: Matrix representation of joint distribution
\item $(p(i)-p_{\opn{draft}}(i))_+$: The positive part of the difference
\item $G_H(t)$: Generating function defined as $\prod_{i\in H}(1+q(i)t)$
\item $W_{n,H}$: Coefficient of $t^n$ in $G_H(t)$
\item $f(x|H)$: The marginal value of element x with respect to set $H$
\item $\opn{Top}_{n-1}(q)$: The top n-1 tokens according to probability in $q$
\item $S \in \mathbb{R}^{\Sigma \times \Sigma^n}$: Variables in the equivalent LP formulation
\item $y \in \mathbb{R}^{\Sigma}, z \in \mathbb{R}^{\Sigma^n}$: Dual variables
\item $\pi_{}(\cdot|j)$: The conditional distribution given $j$
\item $\pi_{i,j}$: Individual elements of the joint distribution matrix
\item $\bar{i}_{2:}$: The sequence $\bar{i}$ with the first element removed
\item $H_i$: Sets constructed by adding elements one by one
\item $\sigma$: An ordering of $\Sigma$ used in the efficient computation algorithm
\item $\operatorname{Coeff}_{t^n}$: Coefficient of $t^n$ in a generating function
\end{itemize}

\section{Additional Experiments}

\begin{table}[ht]
\small
\centering
\setlength{\tabcolsep}{4pt}
\caption{
Average generation length $\tau$ of different MDSD methods
and their ratio $\Delta$ 
on MT-Bench based on Eagle framework. 
}\label{tab:eagle}
\resizebox{1\linewidth}{!}{
\begin{tabular}{lcccccc}
\toprule
 \multirow{2}{*}{Method} & \multicolumn{2}{c}{\# Drafts = 2, \# Steps = 4} & \multicolumn{2}{c}{\# Drafts = 4, \# Steps = 3} & \multicolumn{2}{c}{
EAGLE default sparse tree
 }  \\
\cmidrule(r){2 - 3}\cmidrule(r){4 - 5} \cmidrule(r){6 - 7}
&   $\tau$ & $\Delta$  & $\tau$ & $\Delta$ & $\tau$ & $\Delta$   \\
\midrule
\multicolumn{7}{c}{$T=0.1$} \\
\midrule
  RRS w/ replacement & 3.04 $\pm$ 0.02 & -  & 2.83 $\pm$ 0.02 & - & 3.19 $\pm$ 0.03 & -\\
 RRS w/o replacement & 3.27 $\pm$ 0.02 & 1.07 ($\pm$ 0.02) $\times$& 2.96 $\pm$ 0.02 &  1.05 ($\pm$ 0.01) $\times$  & 3.42 $\pm$ 0.03 & 1.07 ($\pm$ 0.02) $\times$\\
 SpecHub& 3.63 $\pm$ 0.02 & 1.19 ($\pm$ 0.02) $\times$ & - & - & - & -\\
Greedy & 3.62 $\pm$ 0.02  & 1.19 ($\pm$ 0.02) $\times$ & 3.39 $\pm$ 0.01 & 1.20 ($\pm$ 0.02) $\times$   & 3.70 $\pm$ 0.03 & 1.16 ($\pm$ 0.02) $\times$\\
  
\midrule
\multicolumn{7}{c}{$T=0.6$} \\
\midrule
  RRS w/ replacement & 3.22 $\pm$ 0.02 & - & 3.11 $\pm$ 0.01 & - & 3.41 $\pm$ 0.02 & -\\
 RRS w/o replacement & 3.52 $\pm$ 0.02 & 1.09 ($\pm$ 0.02) $\times$ & 3.39 $\pm$ 0.01 & 1.09 ($\pm$ 0.01) $\times$  & 3.71 $\pm$ 0.02 & 1.09 ($\pm$ 0.02) $\times$\\
 SpecHub& 3.52 $\pm$ 0.02 & 1.09 ($\pm$ 0.02) $\times$ & - & - & - & -\\
Greedy & 3.52 $\pm$ 0.02 & 1.09 ($\pm$ 0.02) $\times$ & 3.45 $\pm$ 0.01 & 1.11 ($\pm$ 0.01) $\times$  & 3.66 $\pm$ 0.02  & 1.07 ($\pm$ 0.02) $\times$\\
  
\midrule
\multicolumn{7}{c}{$T=1.0$} \\
\midrule
  RRS w/ replacement & 3.14 $\pm$ 0.02 & - & 3.09 $\pm$ 0.01& - & 3.22 $\pm$ 0.02 & -\\
 RRS w/o replacement & 3.22 $\pm$ 0.02  & 1.02 ($\pm$ 0.02) $\times$  & 3.25 $\pm$ 0.01 & 1.05  ($\pm$ 0.01) $\times$ & 3.43 $\pm$ 0.02 & 1.06 ($\pm$ 0.02) $\times$\\
 SpecHub&  3.35 $\pm$ 0.02 & 1.07 ($\pm$ 0.02) $\times$ &  - & - & - & -\\
Greedy &  3.33 $\pm$ 0.02 & 1.06 ($\pm$ 0.02) $\times$ & 3.34 $\pm$ 0.01 & 1.08 ($\pm$ 0.01) $\times$ & 3.33 $\pm$ 0.02 & 1.03 ($\pm$ 0.02) $\times$\\

\bottomrule
\end{tabular} }
\end{table}

\begin{remark}
For \# Drafts = 2, \# Steps = 4, and $T=0.6$, three methods - RRS without replacement, SpecHub, and Greedy - show similar average generation lengths. After truncating to two decimal places, they appear to be the same. However, they are actually different numbers: 3.51781, 3.51944, 3.51975.
\end{remark}
  
\end{document}